\documentclass[11pt]{article}

\usepackage[final]{acl}
\usepackage{times}
\usepackage{latexsym}
\usepackage[T1]{fontenc}
\usepackage[utf8]{inputenc}
\usepackage{microtype}
\usepackage{inconsolata}
\usepackage{graphicx}
\usepackage{cite}
\usepackage{amsmath,amssymb,amsfonts}
\usepackage{algorithmic}
\usepackage{textcomp}
\usepackage{xcolor}
\usepackage{comment}
\usepackage{multicol}
\usepackage{multirow}
\usepackage{pgfplots} 
\pgfplotsset{compat=1.8}
\usepackage{subcaption}
\usepackage{tikz}
\usepackage{enumitem}

\title{An Exploration of Mamba for Speech Self-Supervised Models}

\author{
  \textbf{Tzu-Quan Lin}\textsuperscript{1,2}$^{\dag}$ \quad
  \textbf{Heng-Cheng Kuo}\textsuperscript{1,3}$^{\dag}$ \quad
  \textbf{Tzu-Chieh Wei}\textsuperscript{1}$^{\ddag}$ \quad
  \textbf{Hsi-Chun Cheng}\textsuperscript{1,2}$^{\ddag}$ \\
  \textbf{Chun Wei Chen}\textsuperscript{1}$^{\ddag}$ \quad
  \textbf{Hsien-Fu Hsiao}\textsuperscript{1}$^{\ddag}$ \quad
  \textbf{Yu Tsao}\textsuperscript{3} \quad
  \textbf{Hung-yi Lee}\textsuperscript{1,2,4} \\
  \textsuperscript{1}National Taiwan University, Taiwan \\
  \textsuperscript{2}Graduate Institute of Communication Engineering, National Taiwan University, Taiwan \\
  \textsuperscript{3}Research Center for Information Technology Innovation, Academia Sinica \\
  \textsuperscript{4}NTU Artificial Intelligence Center of Research Excellence (NTU AI-CoRE) \\
  \texttt{tzuquanlin@gmail.com} \quad \texttt{r11946023@ntu.edu.tw}
}

\newcommand\blfootnote[1]{%
  \begingroup
  \renewcommand\thefootnote{}\footnote{#1}
  \addtocounter{footnote}{-1}%
  \endgroup
}

\begin{document}
\maketitle

\blfootnote{$^\dag$$^\ddag$Equal contribution}

\begin{abstract}
While Mamba has demonstrated strong performance in language modeling, its potential as a speech self-supervised learning (SSL) model remains underexplored, with prior studies limited to isolated tasks. To address this, we explore Mamba-based HuBERT models as alternatives to Transformer-based SSL architectures. Leveraging the linear-time Selective State Space, these models enable fine-tuning on long-context ASR with significantly lower compute. Moreover, they show superior performance when fine-tuned for streaming ASR. Beyond fine-tuning, these models show competitive performance on SUPERB probing benchmarks, particularly in causal settings. Our analysis shows that they yield higher-quality quantized representations and capture speaker-related features more distinctly than Transformer-based models. These findings highlight Mamba-based SSL as a promising and complementary direction for long-sequence modeling, real-time speech modeling, and speech unit extraction. The codebase is available at
\url{https://github.com/hckuo145/Mamba-based-HuBERT}.
\end{abstract}

\section{Introduction}
In recent years, Transformer-based models and their multi-head self-attention mechanisms have achieved remarkable success across various domains~\citep{NIPS2017_3f5ee243, openai2024gpt4technicalreport, devlin-etal-2019-bert, dosovitskiy2021an, radford2022robustspeechrecognitionlargescale}. However, their quadratic computational complexity with respect to sequence length results in high deployment costs. As an alternative, Mamba adopts a Selective State Space architecture that retains content selection capabilities while reducing computational complexity to linear~\citep{gu2024mambalineartimesequencemodeling, DBLP:conf/icml/ZhuL0W0W24, 10890199, lenz2025jamba}. In language modeling tasks, Mamba not only outperforms Transformers of similar scale but also rivals models with twice the number of parameters. Nevertheless, its application in the speech domain has yet to achieve comparable success. Most prior studies evaluate Mamba on a single downstream task, where it usually lags behind Transformer variants~\citep{jiang2025speech, jiang2024dualpathmambashortlongterm, gao2024speech}; those that insert auxiliary blocks can regain accuracy, but probably by forfeiting Mamba’s hallmark linear-time scaling~\citep{li2024spmambastatespacemodelneed}—and none provides a unified, cross-task evaluation. These limitations motivate a systematic exploration of Mamba-based SSL models that can be pretrained on large unlabelled speech and rapidly adapted, through lightweight fine-tuning, to a diverse set of tasks.

In this paper, we draw inspiration from the self-supervised training procedure of HuBERT~\citep{hsu2021hubert} and systematically train Mamba-based self-supervised speech models, termed Mamba-based HuBERT.
The study begins with an in-depth analysis of their characteristics in automatic speech recognition (ASR) tasks, validating two key advantages: faster inference and linear scaling in sequence length. We quantify the MACs per second and real-time factor (RTF) of Mamba-based and Causal Transformer-based HuBERT models as input lengths increase from 5 seconds to 5 minutes. The results show that Mamba’s computational cost remains nearly constant regardless of sequence length, while the computation and memory demands of the Causal Transformer grow rapidly, leading to out-of-memory (OOM) errors beyond 80 seconds.

These findings confirm that the Mamba architecture enables speech SSL models to handle long-context audio effectively, offering a more efficient alternative to Transformers. Therefore, we further fine-tune an External Bidirectional Mamba (ExtBiMamba) model~\citep{zhang2025mambaspeechalternativeselfattention} for long-context ASR, where entire speeches are processed without sentence segmentation. In this setting, the word error rate (WER) is reduced from 13.37\% to 11.08\%, while the Transformer model of the same size fails to run due to memory limitations.

Beyond long-context modeling, Mamba’s inherent causal nature demonstrates potential for streaming ASR. Under the constraint of using only past information, a 78M parameter Mamba-based HuBERT achieves a WER of 15.77\%, outperforming a 94M parameter Causal Transformer-based HuBERT (16.66\%), achieving better accuracy with fewer parameters.

After validating its potential in ASR tasks, we further investigate the generalizability of Mamba-based HuBERT models. Using phone purity~\citep{hsu2021hubert} and canonical correlation analysis (CCA)~\citep{hotelling1936relations}, we analyze the learned representations and find that Mamba-based HuBERT captures phonetic and speaker-related features more distinctly than its Transformer-based counterpart. 
We also evaluate four representative downstream tasks from the SUPERB benchmark~\citep{yang21c_interspeech}. 
In the causal setting, Mamba-based HuBERT surpasses Transformer-based HuBERT in phonetic and speaker-related tasks, while slightly lagging in others, though still achieving higher overall performance. In the bidirectional setting, the small-size ExtBiMamba-based HuBERT outperforms Transformer-based HuBERT on nearly all tasks, while the base-size variant underperforms across the board, indicating room for improvement in scalability.

In conclusion, our contributions are as follows:
\begin{itemize}[noitemsep]
\item We find that Mamba’s inherent causal architecture makes it particularly well-suited for building causal speech SSL models, outperforming its Transformer-based counterpart
\item We show that Mamba-based HuBERT models offer advantages when fine-tuned for long-context ASR and streaming ASR.
\item We find that Mamba-based HuBERT models produce quantized representations with higher phonetic quality, which is beneficial for spoken language models that take SSL units as input.
\item To the best of our knowledge, this is the first comprehensive exploration of Mamba-based HuBERT models as both speech foundation models and feature extractors.
\end{itemize}

\begin{figure*}[ht]
    \centering
  \begin{subfigure}{\linewidth}
    \centering
    \scalebox{0.85}{
    \begin{tikzpicture}
          \definecolor{MyOrange}{HTML}{FF7F0E}
          \definecolor{MyBlue}{HTML}{1F77B4}
          \definecolor{MyLightOrange}{HTML}{FFBF86}
          \definecolor{MyLightBlue}{HTML}{8FBBD9}
        
          \draw[thick, MyOrange] (0,0) -- +(0.8,0);
          \node[MyOrange] at (0.4,0) {\scalebox{2}{\pgfuseplotmark{triangle*}}};
          \node[anchor=west] at (1.4,0) {Mamba};
        
          \draw[thick, MyBlue] (3.4,0) -- +(0.8,0);
          \node[MyBlue] at (3.8,0) {\scalebox{1.5}{\pgfuseplotmark*}};
          \node[anchor=west] at (4.45,0) {Causal Transformer};
        
          \draw[thick, MyLightOrange] (7.8,0) -- +(0.8,0);
          \node[MyLightOrange] at (8.2,0) {\scalebox{2}{\pgfuseplotmark{triangle*}}};
          \node[anchor=west] at (9.0,0) {ExtBiMamba};
        
          \draw[thick, MyLightBlue] (11.6,0) -- +(0.8,0);
          \node[MyLightBlue] at (12.0,0) {\scalebox{1.5}{\pgfuseplotmark*}};
          \node[anchor=west] at (12.8,0) {Transformer};
        \end{tikzpicture}
    }
  \end{subfigure}
      \begin{subfigure}{0.36 \linewidth}
        \centering
        \includegraphics[width=1.0\linewidth]{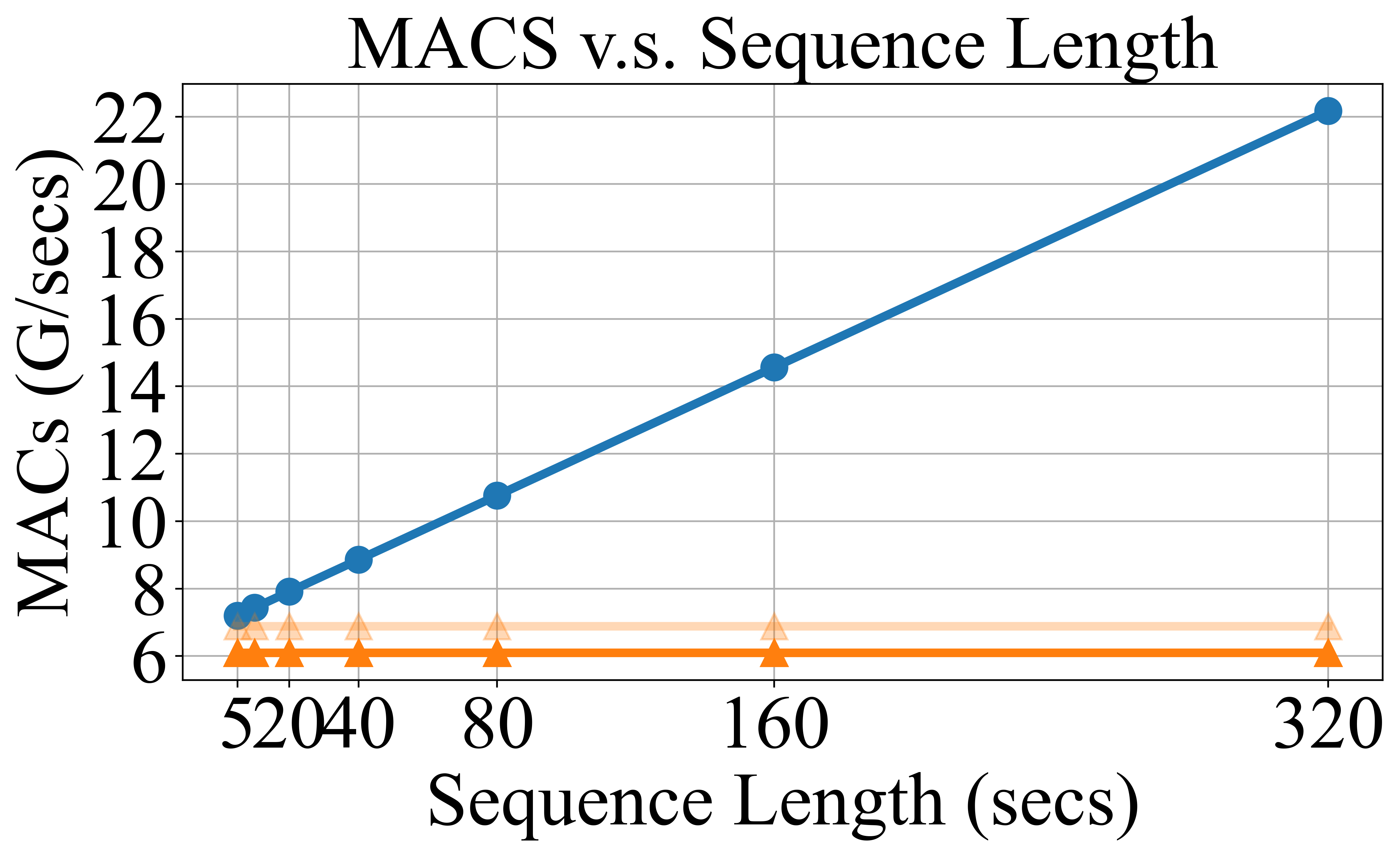}
      \end{subfigure}
      \begin{subfigure}{0.36 \linewidth}
        \centering
        \includegraphics[width=1.0\linewidth]{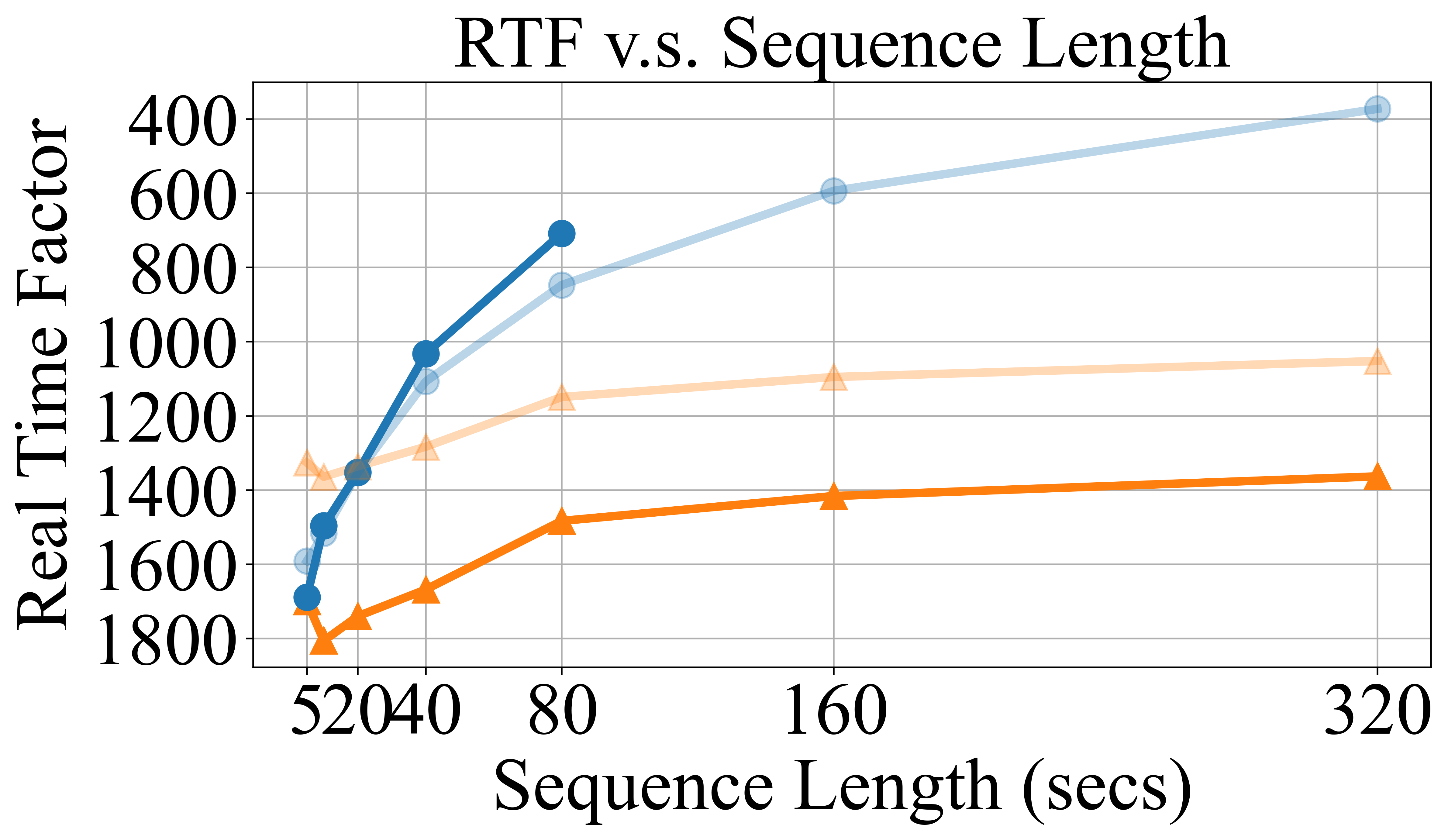}
      \end{subfigure}
    \caption{MACs (G/sec) and Real-Time Factor (RTF) of different HuBERT models at varying sequence lengths.}
    \label{fig:seq-compute-cost}
\end{figure*}

\section{Related Works}
\subsection{Speech Representation Learning and SUPERB}
Over the past few years, self-supervised learning (SSL) has been adopted for speech representation, pretraining models on large amounts of unlabeled audio to extract reusable latent knowledge. Representative methods include wav2vec 2.0~\citep{baevski2020wav2vec20frameworkselfsupervised}, which uses masking and contrastive learning and approaches fully supervised ASR performance with only minutes of labeled data, and HuBERT, which employs k-means pseudo-labels to further improve recognition and generation tasks. These frameworks greatly reduce the need for manual annotation and enable a single model to be reused across speaker identification, emotion recognition, and other downstream tasks.

SUPERB~\citep{yang21c_interspeech} is designed to systematically evaluate the generality and reusability of such pretrained models.
By freezing the pretrained backbone and fine-tuning only lightweight task heads, researchers can evaluate ASR, intent classification, speaker verification, emotion recognition, and more within a unified pipeline, thus focusing on representation quality while lowering experimental overhead.

\subsection{Mamba}
Mamba is a state space model (SSMs) whose discrete-time formulas are expressed as follow:
\begin{align}
     h_t = \overline{A}h_{t-1}+ \overline{B}x_t,\quad y_t = Ch_t
\end{align}
where $h(t)$ is the state vector, $\overline{A}$ is the state transition matrix,  $\overline{B}$ regulates the interaction between input and state, and $C$ maps the state to the output. 

Since $\overline{A}$ and $\overline{B}$ are discrete-time parameters obtained from an underlying continuous system, they are not learned directly through back-propagation. Instead, they are typically approximated via \textbf{Zero-Order Hold (ZOH)} as follows:
\begin{equation}
\begin{aligned}
  \overline{A} &= \exp(\Delta A),\\
  \overline{B} &= (\Delta A)^{-1}\!\left(exp(\Delta A)-I\right)\cdot\!\Delta B
\end{aligned}
\end{equation}
where $A$ and $B$ are the continuous-time equivalents of $\overline{A}$ and $\overline{B}$, and $\Delta$ represents the discretization step. In practice, we parameterize and train the continuous matrices $A$ and $B$. Then, at each forward pass, they are converted to their discrete forms $\overline{A}$ and $\overline{B}$ via ZOH. This transformation allows SSMs to be efficiently applied in discrete-time tasks while preserving their capability to capture long-range dependencies.

To enhance the context-aware capability of SSMs, Gu et al. introduced a \textbf{selection mechanism}, which dynamically adjusts system parameters based on the input signal:
\begin{equation}
\begin{aligned}
    B &= f_B(x),\quad C = f_C(x),\\
    \Delta &= \mathrm{Broadcast}_D\!\left(f_\Delta(x)\right)
\end{aligned}
\end{equation}
where $f_B$, $f_C$, and $f_\Delta$ are parameterized linear projection. Specifically, $f_\Delta(x)$ is a one-dimensional linear projection, and Broadcast$_D$ replicate this value $D$ times to match the shape of $h_t$.
This selection mechanism allows the model to flexibly adapt its state transition and output mapping according to the current input, improving its adaptability for long-sequence modeling.

\subsection{Application of Mamba}
Mamba has been applied to various speech-related tasks. In speech separation and enhancement~\citep{jiang2025speech, jiang2024dualpathmambashortlongterm, li2024spmambastatespacemodelneed, 10832332}, Mamba has been integrated into U-Net architectures to leverage its linear-time complexity for long-sequence modeling while maintaining low computational cost. However, many of these works report only comparable or even slightly worse performance than original Transformer-based baselines, and often require hybrid designs involving both Mamba and Transformer components.

Notably, in causal settings, Mamba-based HuBERT models often achieve the best results among all tested architectures. Mamba has also shown strong potential in streaming ASR~\citep{fang2025mamba}, where its linear computational complexity is well-suited for low-latency applications. By incorporating mechanisms such as lookahead, Unimodal Aggregation (UMA), and Early Termination (ET), Mamba-based streaming models can significantly reduce latency compared to conventional causal Transformer and Conformer models, without sacrificing recognition accuracy.

In the context of self-supervised learning, Zhang et al.~\citep{10889111} utilize Mamba-based SSL models as a tool to analyze Mamba's behavior in other speech processing tasks. 
In contrast, our work investigates the potential of Mamba-based SSL models as a speech foundation model and feature extractor, with comprehensive studies on its fine-tuning performance, probing results, and representational properties.

Moreover, Mamba has been employed in the Self-Supervised Audio Mamba (SSAM) framework to learn general-purpose audio representations~\citep{Yadav24-AM}. SSAM takes advantage of Mamba’s efficient state-space modeling to capture long-range dependencies and produce robust, scalable audio embeddings with both low compute cost and strong contextual modeling. 
However, this work focuses primarily on general audio and does not include experiments specific to speech.

\section{Experiments}
To construct our models, we replace the Transformer blocks in HuBERT with Mamba blocks, while preserving the feature extraction modules.
Specifically, the input waveform is first processed by the standard 7-layer CNN feature encoder (with a 25 ms receptive field and 20 ms frame shift) followed by a convolution-based positional encoder.
The Transformer blocks follow FAIR’s original PyTorch implementation~\citep{paszke2019pytorch}, while the Mamba blocks adopt the implementation provided by Gu \& Dao~\citep{gu2024mambalineartimesequencemodeling}.
By default, the Mamba blocks do not include a feedforward MLP module. If an MLP is added after the Mamba layer, we explicitly denote the model as Mamba+MLP.

We follow the HuBERT Base training pipeline~\citep{hsu2021hubert} to train our Mamba-based HuBERT models. In the first iteration, MFCC features are used as targets, and the model is trained for 250k steps. In the second iteration, we restart training from scratch using the sixth-layer output from the first iteration as targets, and train for another 400k steps.
All pre-training hyperparameters, except for the batch size, follow the default settings of HuBERT. 
Specifically, models are pre-trained on the full 960-hour LibriSpeech dataset. We use the Adam optimizer with a linear warm-up for the first 8\% of training updates, followed by a linear decay scheduler. To ensure stability, the peak learning rate is set to 5e-5 for BiMamba Base models , and 5e-4 for all other configurations.
The reported results are based on a single training run, without averaging across multiple trials.
Due to limited computational resources, we train on a single NVIDIA V100 GPU and adopt a per-GPU batch size that is eight times larger than that used in the original HuBERT paper, which employed 32 GPUs.
As a result, the total audio duration seen per batch is approximately one-fourth of the original setting. For a fair comparison, we apply the same training setup to Transformer-based HuBERT models trained from scratch.

We evaluate models under two settings: causal and bidirectional. The causal setting includes Mamba and causal Transformer models, where the latter uses a standard MHSA mechanism with a lower-triangular mask applied to the attention map. For the bidirectional setting, our experiments include External Bidirectional Mamba (ExtBiMamba), Inner Bidirectional Mamba (InnBiMamba), and the standard Transformer. The definitions of ExtBiMamba and InnBiMamba follow~\citep{zhang2025mambaspeechalternativeselfattention}. 
Both BiMambas adopt a two-branch design: one branch processes the sequence in its original order, while the other processes a time-reversed version; the backward output is reversed again and summed with the forward output. ExtBiMamba realizes these branches as two fully independent Mamba encoder layers with distinct input and output projections, whereas InnBiMamba shares these projections and only duplicates the convolutional and SSM modules. For clarity, we provide an illustration of these differences in Appendix~\ref{sec:bimamba_appendix}.
Based on prior work showing stronger performance, ExtBiMamba is adopted as the default Mamba variant. We further perform an ablation study to evaluate the impact of using InnBiMamba.

\section{Fine-tuning Results on ASR}
\label{sec:mamba_advantages_asr}

We begin by evaluating the performance of fine-tuning Mamba-based HuBERT models for automatic speech recognition (ASR) in both bidirectional and causal settings. 
All fine-tuning setups follow wav2vec2~\citep{baevski2020wav2vec20frameworkselfsupervised} and HuBERT~\citep{hsu2021hubert}.
Before presenting the results, we first highlight the computational efficiency of Mamba-based HuBERT models when handling long sequences.

\subsection{Computational Efficiency Across Sequence Lengths}
\label{subsec:computational_efficiency}

Mamba-based HuBERT models leverage the linear-time complexity of State Space Models with respect to sequence length, offering a computationally efficient alternative to Transformer-based approaches, whose self-attention mechanism incurs quadratic complexity.

To validate this advantage, we measure the Multiply-Accumulate operations (MACs) and Real-Time Factor (RTF)~\citep{feng2023superb, lin2022compressing} across a range of sequence lengths: 5, 10, 20, 40, 80, 160, and 320 seconds. 
Notably, the reported MACs are measured in \textbf{MACs/second}, ensuring that the values directly reflect computational efficiency relative to input duration. 
Each RTF value is averaged over 10 runs to ensure stability, and all RTF measurements are conducted on a single NVIDIA RTX A6000 48 GB GPU.

Figure~\ref{fig:seq-compute-cost} (left panel) shows that the MACs for Mamba-based HuBERT models remain nearly constant across all sequence lengths. In contrast, the MACs for Transformer-based HuBERT models increase sharply with longer sequences, highlighting the computational overhead of attention mechanisms.

A similar trend is observed for RTF (Figure~\ref{fig:seq-compute-cost}, right panel). Although all models exhibit increasing RTF with longer sequences, Mamba-based HuBERT models consistently maintain lower RTF values, particularly for long sequences. 
Notably, Causal Transformer begins to encounter out-of-memory (OOM) errors beyond a sequence length of 80 seconds, due to the additional computation required by the causal attention mask.
This result underscores the superior efficiency of Mamba-based HuBERT models in long-context processing, making it a strong candidate for fine-tuning tasks that require extensive contextual information, such as long-context ASR.

\begin{figure}[ht]
  \centering
  
  \begin{subfigure}{\linewidth}
    \centering
    \scalebox{0.75}{
    \begin{tikzpicture}
          \definecolor{MyOrange}{HTML}{FF7F0E}
          \definecolor{MyBlue}{HTML}{1F77B4}
          \definecolor{MyLightOrange}{HTML}{FFBF86}
          \definecolor{MyLightBlue}{HTML}{8FBBD9}

          \draw[thick, MyOrange] (0,0) -- +(0.8,0);
          \node[MyOrange] at (0.4,0) {\scalebox{2}{\pgfuseplotmark{triangle*}}};
          \node[anchor=west] at (1.0,0) {Mamba};

          \draw[thick, MyBlue] (3.5,0) -- +(0.8,0);
          \node[MyBlue] at (3.9,0) {\scalebox{1.5}{\pgfuseplotmark*}};
          \node[anchor=west] at (4.5,0) {Causal Transformer};

          \draw[thick, MyLightOrange] (0,-0.6) -- +(0.8,0);
          \node[MyLightOrange] at (0.4,-0.6) {\scalebox{2}{\pgfuseplotmark{triangle*}}};
          \node[anchor=west] at (1.0,-0.6) {ExtBiMamba};

          \draw[thick, MyLightBlue] (3.5,-0.6) -- +(0.8,0);
          \node[MyLightBlue] at (3.9,-0.6) {\scalebox{1.5}{\pgfuseplotmark*}};
          \node[anchor=west] at (4.5,-0.6) {Transformer};
    \end{tikzpicture}
    }
  \end{subfigure}

  \vspace{0.5em}

  \begin{subfigure}{0.85\linewidth}
    \centering
    \includegraphics[width=\linewidth]{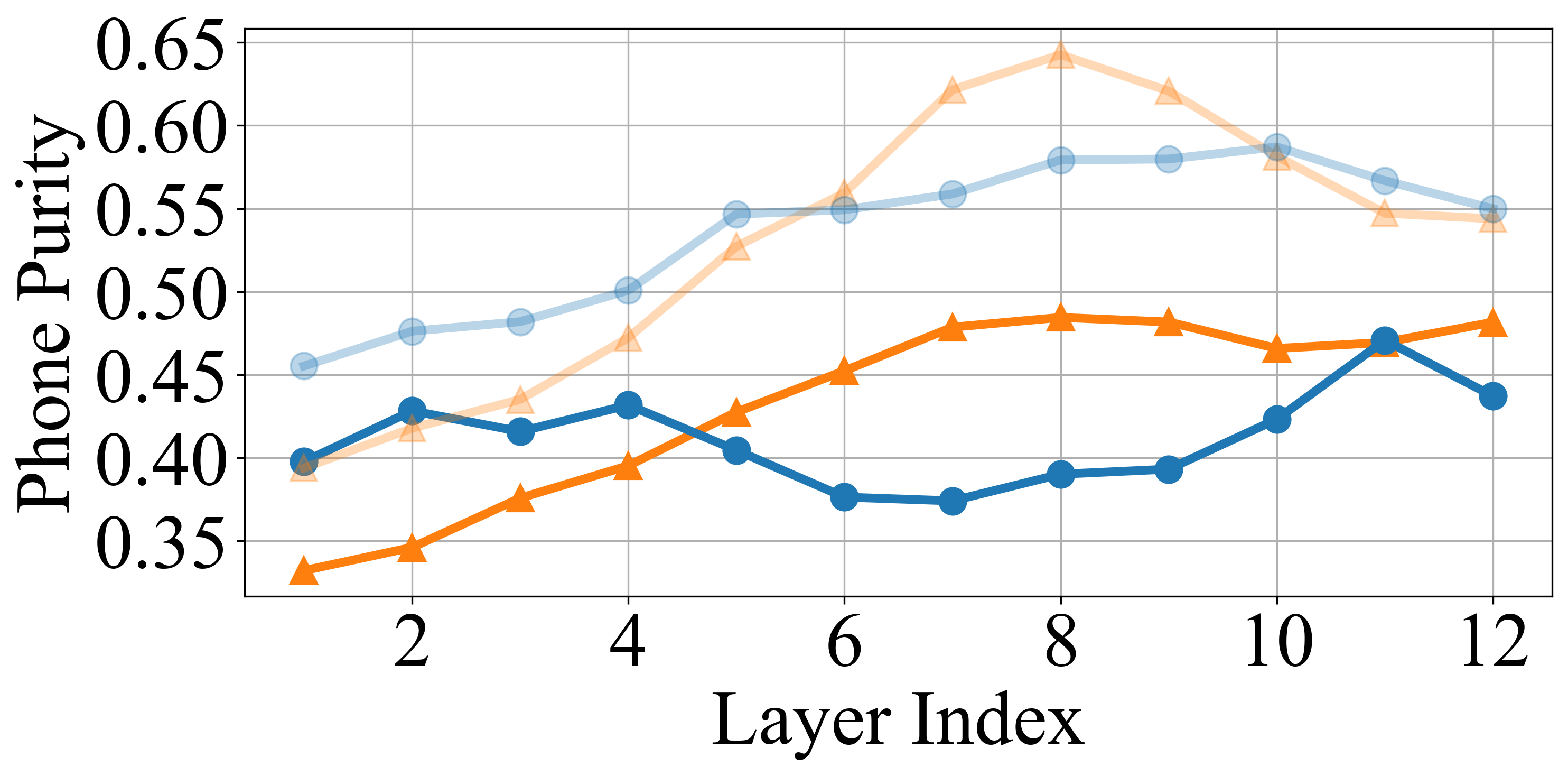}
  \end{subfigure}
  \caption{Layer-wise phone purity of HuBERT models.}
  \label{fig:phone_purity}
\end{figure}
\begin{table}[h]
\centering
\caption{Application of fine-tuning long-context ASR model from Mamba-based HuBERT models, evaluated in terms of WER\%($\downarrow$). Both models contain approximately 94.7M parameters.
}
\scalebox{0.83}{
\renewcommand{\arraystretch}{0.95}
\begin{tabular}{l|cc}
\hline
Model & Utterance level & Document level \\
\hline
\hline
Transformer Base & 11.86 & OOM \\
ExtBiMamba Base & 13.37 & \textbf{11.08} \\
\hline
\end{tabular}
}
\label{tab:long_asr}
\end{table}

\subsection{Fine-tuning for Long-Context ASR}
\label{subsec:long_context_asr}

The efficiency of Mamba-based HuBERT models in handling long sequences makes them particularly well-suited for fine-tuning long-context ASR models, where entire documents are processed as single inputs—unlike conventional utterance-level ASR, which operates on short, segmented speech.
To demonstrate this advantage, we fine-tune a long-context ASR model based on bidirectional Mamba-based HuBERT models.

We conduct experiments on the TEDLIUM3 dataset~\citep{hernandez2018ted}, using its train split for fine-tuning, dev split for validation, and test split for evaluation, with talks longer than 20 minutes removed from all sets.
A 12-layer convolutional encoder is appended to the SSL models, and the entire model is jointly fine-tuned using the CTC loss.

Table~\ref{tab:long_asr} presents the results of fine-tuning different models.
At the utterance level, fine-tuning ExtBiMamba yields a WER of 13.37\%, which is higher than the 11.86\% achieved by fine-tuning Transformer.
However, when ExtBiMamba is fine-tuned and evaluated on full documents, its WER significantly improves to 11.08\%, demonstrating the benefits of leveraging broader context.
Notably, Transformer-based HuBERT models fail to process document-length inputs due to out-of-memory (OOM) errors, further underscoring Mamba’s practical advantage.

To further understand the benefits of long-context ASR, we performed paired t-tests comparing WER across different experimental settings. For ExtBiMamba, document-level outperformed utterance-level with p-value = 0.001. These results demonstrate that Mamba's performance improves significantly when leveraging full document context and confirm statistically significant differences between the two approaches.
Qualitatively, long-context ASR achieves more consistent recognition of rare or challenging words, particularly those that appear multiple times across a document. In contrast, utterance-level ASR often produces inconsistent transcriptions for the same term, highlighting the value of global context in long-context processing.

\begin{table}[h]
    \centering
    \caption{Application of fine-tuning causal ASR model from Mamba-based HuBERT models.}
    \scalebox{1.0}{
    \renewcommand{\arraystretch}{0.95}
        \begin{tabular}{l|c c}
        \hline
        \multirow{2}{*}{Model} & Parameters & ASR \\ 
        & M & WER\%$\downarrow$ \\
        \hline
        \hline
        Causal Trans. Base & 94.7 & 16.66 \\
        Mamba Base & 78.2 & \textbf{15.77} \\
        \cline{1-3}
        \end{tabular}
    }
    \label{tab:causal-asr}
\end{table}
\begin{figure*}[ht]
    \centering
    
    \begin{subfigure}{\linewidth}
        \centering
        \scalebox{0.8}{
        \begin{tikzpicture}
          \definecolor{MyOrange}{HTML}{FF7F0E}
          \definecolor{MyBlue}{HTML}{1F77B4}
          \definecolor{MyLightOrange}{HTML}{FFBF86}
          \definecolor{MyLightBlue}{HTML}{8FBBD9}
        
          \draw[thick, MyOrange] (0,0) -- +(0.8,0);
          \node[MyOrange] at (0.4,0) {\scalebox{2}{\pgfuseplotmark{triangle*}}};
          \node[anchor=west] at (1.4,0) {Mamba};
        
          \draw[thick, MyBlue] (3.4,0) -- +(0.8,0);
          \node[MyBlue] at (3.8,0) {\scalebox{1.5}{\pgfuseplotmark*}};
          \node[anchor=west] at (4.45,0) {Causal Transformer};
        
          \draw[thick, MyLightOrange] (7.8,0) -- +(0.8,0);
          \node[MyLightOrange] at (8.2,0) {\scalebox{2}{\pgfuseplotmark{triangle*}}};
          \node[anchor=west] at (9.0,0) {ExtBiMamba};
        
          \draw[thick, MyLightBlue] (11.6,0) -- +(0.8,0);
          \node[MyLightBlue] at (12.0,0) {\scalebox{1.5}{\pgfuseplotmark*}};
          \node[anchor=west] at (12.8,0) {Transformer};
        \end{tikzpicture}
        }
    \end{subfigure}

    \vspace{0.5em}
    
    \begin{subfigure}{0.32\linewidth}
        \centering
        \includegraphics[width=\linewidth]{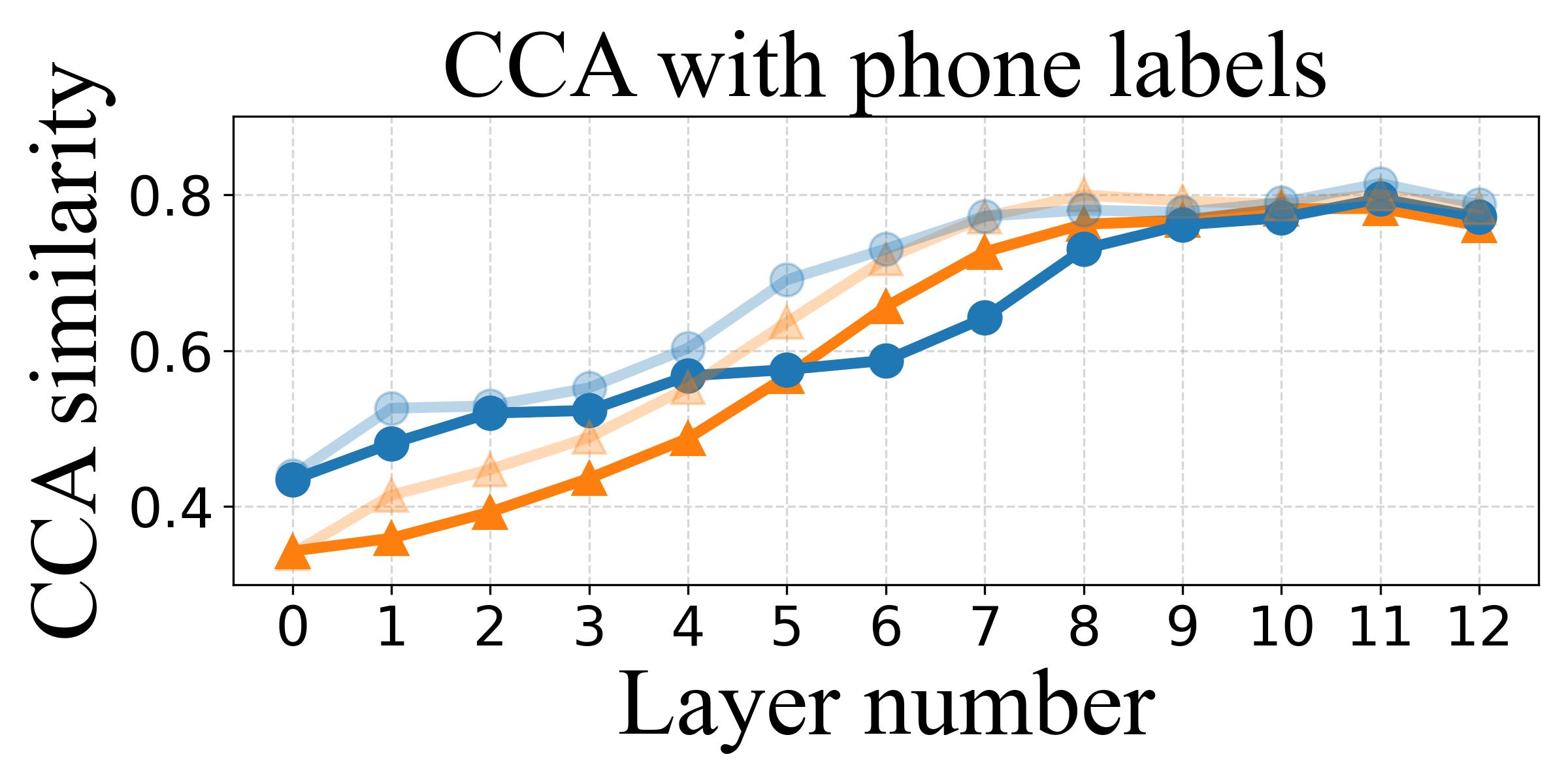}
    \end{subfigure}
    \hfill
    \begin{subfigure}{0.32\linewidth}
        \centering
        \includegraphics[width=\linewidth]{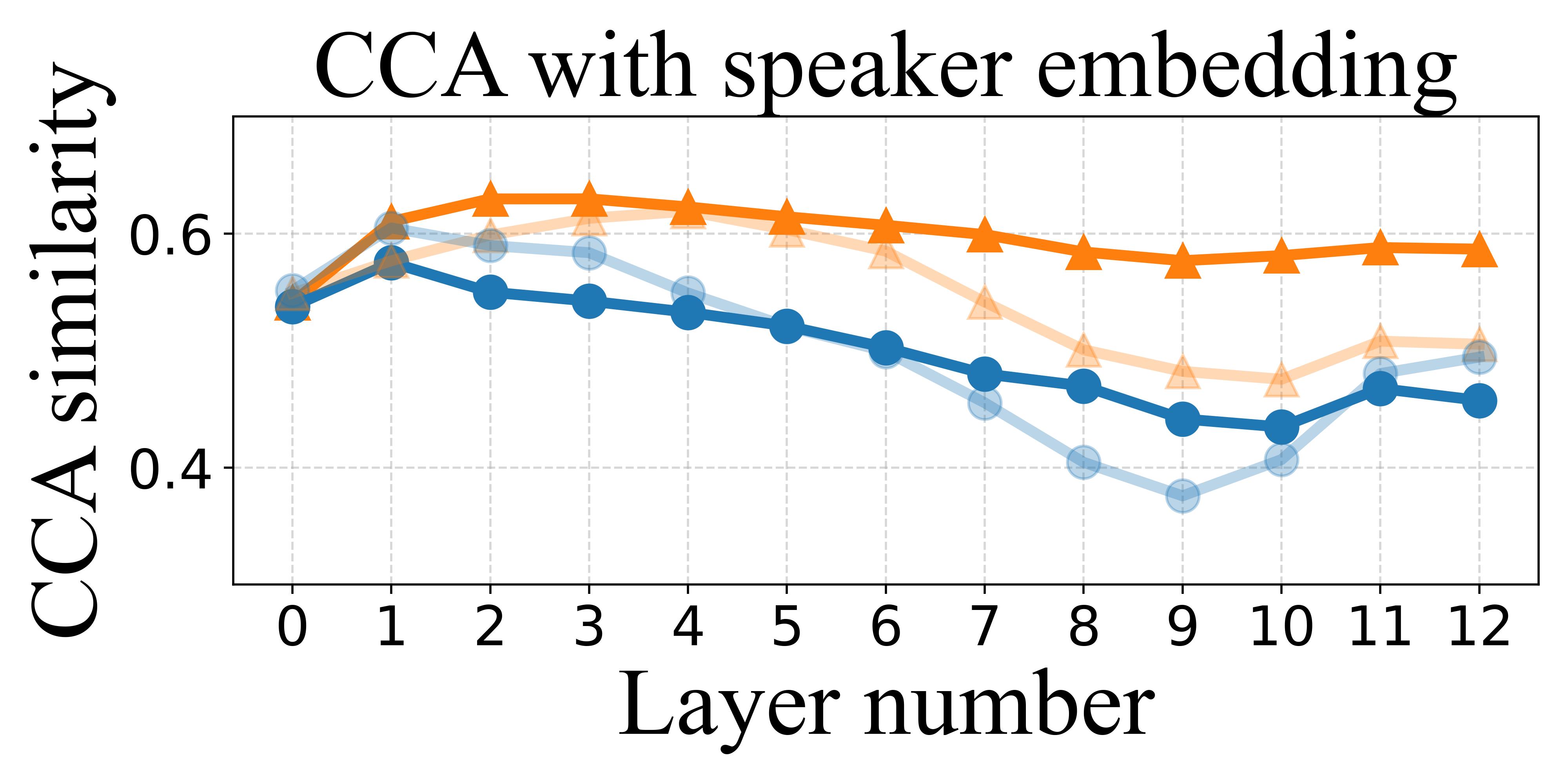}
    \end{subfigure}
    \hfill
    \begin{subfigure}{0.32\linewidth}
        \centering
        \includegraphics[width=\linewidth]{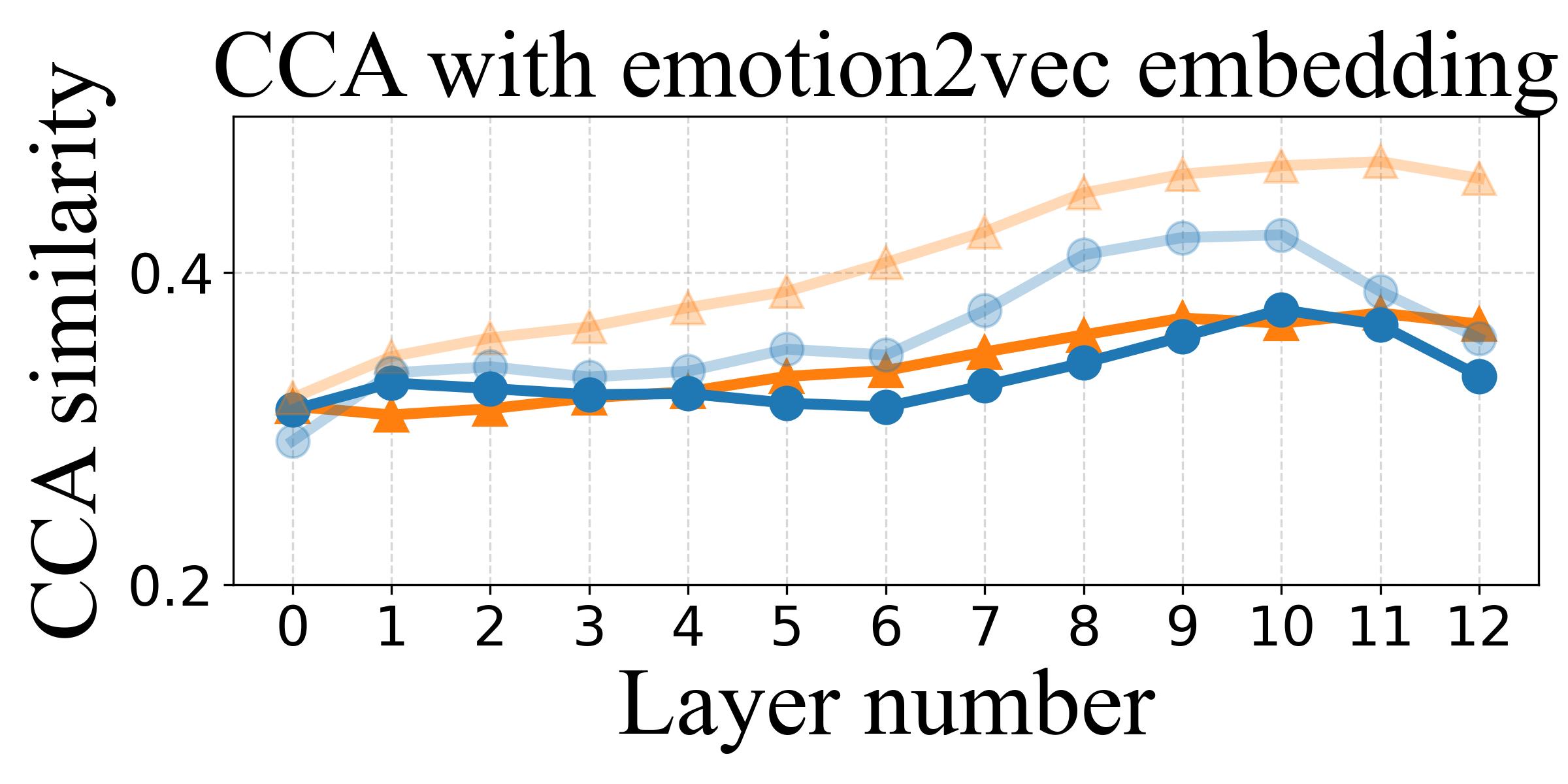}
    \end{subfigure}

    \caption{Layer-wise analysis results for different HuBERT models. Each plot below shows the CCA similarity to different label types: phone labels, speaker embedding, and emotion embedding.}
    \label{fig:layerwise}
\end{figure*}

\subsection{Fine-tuning for Causal ASR}
\label{subsec:causal_asr}

We further examine another scenario of fine-tuning an ASR model: causal ASR, where the causal speech SSL models are fine-tuned to predict the current token using only past information. 
For this task, we use the LibriSpeech 100-hour set for training and the test-clean set for evaluation.
This setting is critical for streaming ASR, where real-time transcription is required.

Table~\ref{tab:causal-asr} presents the results of fine-tuning SSL on causal ASR, where no lookahead is employed for all models. Mamba Base (78.2M parameters) achieves a word error rate (WER) of 15.77\%, outperforming the Causal Transformer Base (94.7M parameters) with a WER of 16.66\%. This result is notable because Mamba achieves superior ASR performance with approximately 17\% fewer parameters. 

\section{In-depth Analysis of Learned Representation}
\label{sec:in_depth_representation_analysis}

Following the exploration of fine-tuning, this section delves deeper into the characteristics of the learned representations.

\subsection{Phone Purity of Quantized Representations}
\label{subsec:phone_purity_revised}
The quality of quantized SSL representations is crucial for various applications, for example, the input token of spoken language modeling~\citep{lakhotia2021generative, hassid2023textually, arora2025landscape}. 
To evaluate the quality of quantized representations, we extract features from different layers of the SSL models, perform k-means clustering, and then calculate the phone purity of these clusters with respect to ground-truth phone labels.
The phone purity is computed following the definition provided in the HuBERT paper~\citep{hsu2021hubert}, and can be regarded as a measure of how well the quantized representations correlate with underlying phonemes.

Figure~\ref{fig:phone_purity}
illustrates the phone purity across different layers for Mamba-based and Transformer-based HuBERT models.
Except for Causal Transformer, phone purity typically increases from the initial layers, reaches a peak in the mid-to-late layers, and then may slightly decline towards the final layers. 

In the causal setting, Mamba consistently demonstrates higher peak phone purity than the Causal Transformer, possibly indicating better encoding of phonetic information.
In the bidirectional setting, an interesting phenomenon emerges: while ExtBiMamba performs worse than the Transformer in utterance-level ASR fine-tuning, its peak phone purity slightly surpasses that of the Transformer in certain mid-to-late layers.
This suggests that the quality of quantized representations does not necessarily align with fine-tuning performance.

We report results with $k=100$ for the k-means clustering here, while additional results with $k=500$ and $k=1000$ are provided in Appendix~\ref{sec:phone_purity_additional}, showing overall trends consistent with the $k=100$ case.

\begin{table*}
    \centering
     \caption{SUPERB probing results for causal speech self-supervised models.}
    \scalebox{0.9}{
    \renewcommand{\arraystretch}{0.95}
    \begin{tabular}{llc|ccccc}
    \hline
    \multirow{2}{*}{Size} & \multirow{2}{*}{Model} & Parameters & PR & SID & ER & IC & \multirow{2}{*}{$\text{SUPERB}_{S}$}\\ 
    & & M & PER\%$\downarrow$ & ACC\%$\uparrow$ & ACC\%$\uparrow$ & ACC\%$\uparrow$ & \\
    \hline
    \hline
    \multirow{4}{*}{Base} & Causal Transformer & 94.7 & 13.87 & 60.04 & \textbf{63.33} & \textbf{94.23} & 805.44 \\ 
    & Transformer + Causal Mask & 94.7 & 25.57 & \textbf{74.57} & 58.77 & 80.07 & 734.68 \\
    & Mamba + MLP & 94.7 & 11.72 & 73.48 & 61.72 & 89.62 & \textbf{823.15} \\
    & Mamba & 78.2 & \textbf{11.68} & 73.07 & 61.98 & 87.45 & 817.94 \\
    \hline
    \hline
    \multirow{4}{*}{Small} & Causal Transformer & 23.5 & 15.91 & 58.27 & \textbf{62.07} & \textbf{87.92} & 767.70 \\ 
    & Transformer + Causal Mask & 23.5 & 24.05 & 66.61 & 60.45 & 81.86 & 735.49 \\
    & Mamba + MLP & 33.0 & \textbf{15.28} & \textbf{68.42} & 61.63 & 82.39 & \textbf{777.68} \\
    & Mamba & 23.5 & 17.34 & 66.06 & 60.82 & 85.13 & 766.94 \\ 
    \cline{1-8}
    \end{tabular}
    }
    \label{tab:causal-result}
\end{table*}

\subsection{Similarity of Representations with Speech Attributes}
\label{subsec:cca_analysis_revised}

Layer-wise analysis~\citep{pasad2021layer, lin2024daisy, pasad2024self, lin2024property, lin2025identifying} is a common approach to examine how speech attributes are distributed across different layers of speech SSL models. 
To better understand the property of the learned representations, we use Canonical Correlation Analysis (CCA)~\citep{hotelling1936relations} to assess the similarity between representations across different layers and various speech attributes, including phoneme labels, speaker embeddings, and emotion embeddings. 
For phoneme labels, we convert them into one-hot vectors and compute CCA with mean-pooled phone-level representations. For speaker and emotion embeddings, we compute CCA using mean-pooled representations over the time axis.

Figure~\ref{fig:layerwise} presents the CCA similarity scores.
For CCA with phone labels (left panel), all models show a clear trend where similarity increases with layer depth, peaking at the final or near-final layers. 
Interestingly, while achieving similarly high peak similarity, Mamba-based HuBERT models exhibit a steeper increase across layers, starting from lower similarity in early layers.

For CCA with Resemblyzer speaker embeddings~\citep{wan2018generalized} (middle panel), all models exhibit higher similarity in the early layers.
Notably, Mamba-based HuBERT models generally maintain higher similarity with speaker embeddings than Transformer-based models, indicating that they capture speaker-related features more distinctly.
We have also computed similarity with speaker embeddings from TitaNet~\citep{koluguri2022titanet} and ECAPA-TDNN~\citep{desplanques2020ecapa}; the trends are mostly similar to those observed with Resemblyzer.

For CCA with emotion2vec embeddings~\citep{ma-etal-2024-emotion2vec} (right panel), CCA similarity increases towards the mid-to-late layers for all models. 
ExtBiMamba and Transformer tend to show higher similarity scores compared to the causal models. This suggests that emotion-related information might be captured more effectively with access to bidirectional context.
Compared to Transformer-based models, Mamba-based HuBERT models exhibit a steadily increasing CCA similarity across layers, without a noticeable decline in the final layers.

\section{Downstream Performance Comparison}
\label{sec:comparative_performance}

After evaluating the fine-tuning performance and representation characteristics, we further conduct downstream probing experiments on SUPERB~\citep{yang21c_interspeech}. MiniSUPERB~\citep{wang2023minisuperb} demonstrates that evaluating a subset of SUPERB tasks can sufficiently reflect the ranking observed in the full evaluation. Hence, we select four tasks for evaluation: phoneme recognition (PR), speaker identification (SID), emotion recognition (ER), and intent classification (IC). 
Additionally, we select several representative models for a full SUPERB evaluation to provide a more comprehensive comparison, with results presented in Appendix E.
We follow the default evaluation settings of SUPERB.
Following prior works~\citep{feng2023superb, shi23g_interspeech, shi2023findings, shi2024multiresolution, wang2023_interspeech, lin2025speechft}, we report the widely used overall SUPERB score (SUPERB$_{S}$) computed from these four tasks for easier comparison (see Appendix~\ref{sec:superb-score} for details on the score computation).

\subsection{Probing Results in Causal Setting}
\label{subsec:causal_performance}

Table~\ref{tab:causal-result} presents the performance in the causal setting. 
In the Base-size category, Mamba-based HuBERT models demonstrate strong performance. Mamba + MLP (94.7M parameters) achieves the highest SUPERB$_{S}$ score (823.15) and a competitive PR of 11.72\%. The standard Mamba (78.2M parameters) also performs well, with a PR of 11.68\% and a SUPERB$_{S}$ score of 817.94. Both Mamba variants outperform Causal Transformer (94.7M parameters), which records a PR of 13.87\% and a SUPERB$_{S}$ score of 805.44. While Causal Transformer excels in ER (63.33\%) and IC (94.23\%), Mamba-based HuBERT models consistently achieve better phoneme recognition and overall representation quality.

For Small-size models, Mamba + MLP (33.0M parameters) achieves the highest SUPERB$_{S}$ score (777.68), along with strong PR (15.28\%) and SID (68.42\%). The standard Mamba (23.5M parameters) performs comparably to the Causal Transformer (23.5M parameters), with similar SUPERB$_{S}$ scores (766.94 vs. 767.70). This suggests that even at smaller scales, Mamba-based HuBERT models maintain competitive performance.

In addition, we include a naive baseline, Transformer + Causal Mask, which directly applies a causal mask to a Transformer trained in a bidirectional setting. However, this approach performs significantly worse than the Causal Transformer, highlighting the importance of pre-training with causal behavior.

The probing results on SUPERB are consistent with our observations on phone purity in Section~\ref{subsec:phone_purity_revised}, where Mamba's representations demonstrate superior phonetic information compared to those of causal Transformer.

\begin{table*}
    \centering
     \caption{SUPERB probing results for bidirectional speech self-supervised models.}
    \scalebox{1.0}{
    \renewcommand{\arraystretch}{0.95}
    \begin{tabular}{llc|ccccc}
    \hline
    \multirow{2}{*}{Size} & \multirow{2}{*}{Model} & Parameters & PR & SID & ER & IC & \multirow{2}{*}{$\text{SUPERB}_{S}$} \\
    & & M & PER\%$\downarrow$ & ACC\%$\uparrow$ & ACC\%$\uparrow$ & ACC\%$\uparrow$ & \\
    \hline
    \multirow{2}{*}{Base} & Transformer & 94.7 & \textbf{7.49} & \textbf{75.77} & \textbf{62.36} & \textbf{97.44} & \textbf{868.93} \\
    & ExtBiMamba & 94.3 & 10.65 & 68.31 & 61.24 & 91.7 & 815.38 \\
    \hline
    \hline
    \multirow{2}{*}{Small} & Transformer & 23.5 & 12.21 & 67.29 & 60.46 & \textbf{91.83} & 802.63 \\
    & ExtBiMamba & 23.2 & \textbf{11.38} & \textbf{69.22} & \textbf{62.34} & 86.69 & \textbf{809.18} \\
    \cline{1-8}
    \end{tabular}
    }
    \label{tab:bidirectional-result}
\end{table*}
\begin{table*}
    \centering
    \caption{Ablation on bidirectional Mamba architectures. The definitions of ExtBiMamba and InnBiMamba follow those in prior work~\citep{zhang2025mambaspeechalternativeselfattention}.}
    \scalebox{1.0}{
    \renewcommand{\arraystretch}{0.95}
    \begin{tabular}{lll|ccccc}
    \hline
    \multirow{2}{*}{Size} & \multirow{2}{*}{Model} & Parameters & PR & SID & ER & IC & \multirow{2}{*}{$\text{SUPERB}_{S}$} \\
    & & M & PER\%$\downarrow$ & ACC\%$\uparrow$ & ACC\%$\uparrow$ & ACC\%$\uparrow$ & \\
    \hline
    \multirow{3}{*}{Base} & ExtBiMamba & 94.3 & 10.65 & 68.31 & 61.24 & 91.7 & 815.38 \\
    & InnBiMamba + MLP & 94.7 & \textbf{9.00} & \textbf{70.67} & 61.07 & \textbf{91.56} & 825.17 \\
    & InnBiMamba & 82.9 & 9.62 & 70.79 & \textbf{62.36} & 91.43 & \textbf{832.31} \\
    \hline
    \hline
    \multirow{2}{*}{Small} & ExtBiMamba & 23.2 & \textbf{11.38} & \textbf{69.22} & \textbf{62.34} & 86.69 & \textbf{809.18} \\
    & InnBiMamba & 25.1 & 14.44 & 60.77 & 60.84 & \textbf{87.00} & 767.60 \\
    \cline{1-8}
    \end{tabular}
    }
    \label{tab:innbimamba}
\end{table*}
\subsection{Probing Results in Bidirectional Setting}
\label{subsec:bidirectional_performance}

Table~\ref{tab:bidirectional-result} compares the performance in the bidirectional setting.
For Base-size models, Transformer (94.7M parameters) consistently outperforms ExtBiMamba (94.3M parameters) across all metrics. The Transformer achieves a PR of 7.49\% and a SUPERB$_{S}$ score of 868.93, while ExtBiMamba records a PR of 10.65\% and a SUPERB$_{S}$ score of 815.38. 

In contrast, the Small-size models exhibit a different trend. ExtBiMamba (23.2M parameters) surpasses the Transformer (23.5M parameters) in several metrics: PR (11.38\% vs. 12.21\%), SID (69.22\% vs. 67.29\%), ER (62.34\% vs. 60.46\%), and SUPERB$_{S}$ score (809.18 vs. 802.63). Although the Transformer retains an advantage in IC (91.83\% vs. 86.69\%), ExtBiMamba shows a clear edge in most tasks, indicating its effectiveness at smaller scales.

In summary, the probing performance of Mamba-based HuBERT models in the bidirectional setting is scale-dependent. ExtBiMamba underperforms the Transformer at the base scale but demonstrates a competitive advantage at the small scale. This might indicate that the scalability of Mamba in the bidirectional setting remains a challenge.

Interestingly, despite having higher phone purity as shown in Section~\ref{subsec:phone_purity_revised}, ExtBiMamba does not outperform Transformer in PR probing results.
We attribute this to the fact that high phone purity reflects strong intra-class consistency (frames of the same phone falling into dominant clusters) but does not guarantee the inter-class discriminability required for a linear probe to separate different phonemes. Furthermore, the ranking of SSL models can be influenced by the architecture of downstream heads~\citep{zaiem2023speech}, and disentanglement of specific attributes does not always correlate with downstream performance~\citep{plachouras2025unified}.

\subsection{Ablation on Bidirectional Mamba Architectures}
\label{subsec:bimamba_implementations_new_section}

In response to the initial results, we investigated alternative bidirectional Mamba designs, specifically comparing ExtBiMamba and InnBiMamba. The results are summarized in Table~\ref{tab:innbimamba}.

For Base-size models, InnBiMamba outperforms ExtBiMamba. Notably, InnBiMamba + MLP (94.7M parameters) achieves a PR of 9.00\% and a SUPERB$_{S}$ score of 825.17, surpassing ExtBiMamba (94.3M parameters, PR 10.65\%, SUPERB$_{S}$ 815.38). Even the standard InnBiMamba (82.9M parameters) performs exceptionally well, with a PR of 9.62\% and the highest SUPERB$_{S}$ score of 832.31. It also excels in SID (70.79\%) and ER (62.36\%). 

In contrast, for Small-size models, ExtBiMamba maintains an advantage. 
It achieves a PR of 11.38\% and a SUPERB$_{S}$ score of 809.18 with 23.2M parameters, outperforming InnBiMamba (25.1M parameters, PR 14.44\%, SUPERB$_{S}$ 767.60). 

These results demonstrate that the choice of bidirectional Mamba architecture may significantly impact the performance of speech SSL model, and the optimal design may vary with model size. InnBiMamba shows strong potential at larger scales, while ExtBiMamba is more effective at smaller scales.

We hypothesize that the performance gap in ExtBiMamba Base stems from training instability, a phenomenon also reported in large-scale Mamba-based visual backbones~\citep{suleman2024stablemamba, shaker2025groupmamba, patro2024simba}.
Detailed analysis of our training dynamics supports this hypothesis (see Appendix~\ref{sec:training-stability}).

\section{Conclusion}
In this work, we conduct a comprehensive exploration of Mamba-based HuBERT models, including fine-tuning, representational analyses, and downstream probing.
Our results demonstrate that Mamba-based HuBERT models offer advantages in long-context modeling and real-time recognition.
Additionally, they can extract better quantized speech representations, which are beneficial for certain applications such as spoken language modeling.
In downstream probing, Mamba-based HuBERT models generally perform better in the causal setting.
However, we also observe that they suffer from limited scalability in the bidirectional setting.
We believe this work provides empirical evidence and design guidance for the development of Mamba-based speech self-supervised models.

\section{Limitations}
This study is designed with an emphasis on reproducibility and interpretability, so we adopt relatively conservative and controlled settings for training scale and evaluation scope to clearly attribute model behavior and conclusions. 
Specifically, each pre-training model is trained on a single GPU. Although we maximize the amount of audio to fit within the memory limit of one GPU, the overall batch size remains smaller than in the original HuBERT setup. 
As supporting evidence, our Transformer-based HuBERT trained under this setting achieves lower SUPERB scores compared to the official FAIR-released model. 

This choice makes the study reproducible and cost-efficient, though it may not fully exploit larger-scale regimes regarding model parameters, batch sizes, or diverse training corpora; future work can explore scaling laws and alternative curricula on cluster-level resources to verify if findings such as the bidirectional scalability challenges hold at larger scales.

Efficiency results are reported on a consistent single-GPU setup to ensure stable comparisons; while absolute numbers may vary with hardware or low-level optimizations (kernels, computation graphs), relative trends should remain valid. In long-document scenarios, standard Transformer baselines reached memory limits under our budget, reflecting the well-known quadratic scaling property; with memory optimizations, the absolute processable length could be extended for both models. 
These boundaries reflect the scope of our experimental design and, while they may limit certain aspects, they are not expected to alter the central conclusions of this study.

\section{Acknowledgments}
We thank the National Center for High-performance Computing (NCHC) of the National Institutes of Applied Research (NIAR) in Taiwan for providing computational and storage resources. Additionally, this work was supported by the Ministry of Education (MOE) of Taiwan under the project Taiwan Centers of Excellence in Artificial Intelligence, through the NTU Artificial Intelligence Center of Research Excellence (NTU AI-CoRE).

\bibliography{reference}

@inproceedings{Yadav24-AM,
  title     = {{Audio Mamba: Selective State Spaces for Self-Supervised Audio Representations}},
  author    = {Yadav, Sarthak and Tan, Zheng-Hua},
  year      = {2024},
  booktitle = {Interspeech},
  pages     = {552--556}
}

@inproceedings{pasad2021layer,
  title={{Layer-wise analysis of a self-supervised speech representation model}},
  author={Pasad, Ankita and Chou, Ju-Chieh and Livescu, Karen},
  booktitle={2021 IEEE Automatic Speech Recognition and Understanding Workshop (ASRU)},
  pages={914--921},
  year={2021},
  organization={IEEE}
}

@inproceedings{wan2018generalized,
  title={{Generalized end-to-end loss for speaker verification}},
  author={Wan, Li and Wang, Quan and Papir, Alan and Moreno, Ignacio Lopez},
  booktitle={Proceedings of the IEEE International Conference on Acoustics, Speech, and Signal Processing (ICASSP)},
  pages={4879--4883},
  year={2018},
  organization={IEEE}
}

@article{lin2022compressing,
  title={{Compressing transformer-based self-supervised models for speech processing}},
  author={Lin, Tzu-Quan and Yang, Tsung-Huan and Chang, Chun-Yao and Chen, Kuang-Ming and Feng, Tzu-hsun and Lee, Hung-yi and Tang, Hao},
  journal={arXiv preprint arXiv:2211.09949},
  year={2022}
}

@inproceedings{feng2023superb,
  title={{Superb@ slt 2022: Challenge on generalization and efficiency of self-supervised speech representation learning}},
  author={Feng, Tzu-hsun and Dong, Annie and Yeh, Ching-Feng and Yang, Shu-wen and Lin, Tzu-Quan and Shi, Jiatong and Chang, Kai-Wei and Huang, Zili and Wu, Haibin and Chang, Xuankai and others},
  booktitle={2022 IEEE Spoken Language Technology Workshop (SLT)},
  pages={1096--1103},
  year={2023},
  organization={IEEE}
}

@inproceedings{jiang2025speech,
  title={{Speech slytherin: Examining the performance and efficiency of mamba for speech separation, recognition, and synthesis}},
  author={Jiang, Xilin and Li, Yinghao Aaron and Florea, Adrian Nicolas and Han, Cong and Mesgarani, Nima},
  booktitle={Proceedings of the IEEE International Conference on Acoustics, Speech, and Signal Processing (ICASSP)},
  pages={1--5},
  year={2025},
  organization={IEEE}
}

@inproceedings{fang2025mamba,
  title={{Mamba for Streaming ASR Combined with Unimodal Aggregation}},
  author={Fang, Ying and Li, Xiaofei},
  booktitle={Proceedings of the IEEE International Conference on Acoustics, Speech, and Signal Processing (ICASSP)},
  pages={1--5},
  year={2025},
  organization={IEEE}
}

@inproceedings{yang21c_interspeech,
  title     = {{SUPERB: Speech Processing Universal PERformance Benchmark}},
  author    = {Shu-Wen Yang and Po-Han Chi and Yung-Sung Chuang and Cheng-I Jeff Lai and Kushal Lakhotia and Yist Y. Lin and Andy T. Liu and Jiatong Shi and Xuankai Chang and Guan-Ting Lin and Tzu-Hsien Huang and Wei-Cheng Tseng and Ko-tik Lee and Da-Rong Liu and Zili Huang and Shuyan Dong and Shang-Wen Li and Shinji Watanabe and Abdelrahman Mohamed and Hung-yi Lee},
  year      = {2021},
  booktitle = {Interspeech},
  pages     = {1194--1198}
}

@inproceedings{plachouras2025unified,
  title={{Towards a Unified Representation Evaluation Framework Beyond Downstream Tasks}},
  author={Plachouras, Christos and Guinot, Julien and Fazekas, Gy{\"o}rgy and Quinton, Elio and Benetos, Emmanouil and Pauwels, Johan},
  booktitle={Proceedings of the International Joint Conference on Neural Networks (IJCNN)},
  year={2025},
  organization={IEEE}
}

@article{arora2025landscape,
  title={{On the landscape of spoken language models: A comprehensive survey}},
  author={Arora, Siddhant and Chang, Kai-Wei and Chien, Chung-Ming and Peng, Yifan and Wu, Haibin and Adi, Yossi and Dupoux, Emmanuel and Lee, Hung-Yi and Livescu, Karen and Watanabe, Shinji},
  journal={arXiv preprint arXiv:2504.08528},
  year={2025}
}

@article{hsu2021hubert,
  title={{HuBERT: Self-supervised speech representation learning by masked prediction of hidden units}},
  author={Hsu, Wei-Ning and Bolte, Benjamin and Tsai, Yao-Hung Hubert and Lakhotia, Kushal and Salakhutdinov, Ruslan and Mohamed, Abdelrahman},
  journal={IEEE/ACM Transactions on Audio, Speech, and Language Processing},
  volume={29},
  pages={3451--3460},
  year={2021},
  publisher={IEEE}
}

@article{hotelling1936relations,
  title={{Relations between two sets of variates}},
  author={Hotelling, Harold},
  journal={Biometrika},
  volume={28},
  number={3/4},
  pages={321--377},
  year={1936},
  publisher={Oxford University Press}
}

@inproceedings{ma-etal-2024-emotion2vec,
  title = {{emotion2vec: Self-Supervised Pre-Training for Speech Emotion Representation}},
  author = {Ma, Ziyang and Zheng, Zhisheng and Ye, Jiaxin and Li, Jinchao and Gao, Zhifu and Zhang, ShiLiang and Chen, Xie},
  booktitle = {Findings of the Association for Computational Linguistics: ACL 2024},
  pages = {15747--15760},
  year = {2024},
  publisher = {Association for Computational Linguistics}
}

@inproceedings{hernandez2018ted,
  title={{TED-LIUM 3: Twice as much data and corpus repartition for experiments on speaker adaptation}},
  author={Hernandez, Fran{\c{c}}ois and Nguyen, Vincent and Ghannay, Sahar and Tomashenko, Natalia and Esteve, Yannick},
  booktitle={Proceedings of the 20th International Conference on Speech and Computer (SPECOM)},
  pages={198--208},
  year={2018},
  organization={Springer}
}

@inproceedings{zaiem2023speech,
  title     = {{Speech Self-Supervised Representation Benchmarking: Are We Doing it Right?}},
  author    = {Salah Zaiem and Youcef Kemiche and Titouan Parcollet and Slim Essid and Mirco Ravanelli},
  year      = {2023},
  booktitle = {Interspeech},
  pages     = {2873--2877}
}

@inproceedings{hassid2023textually,
 author = {Hassid, Michael and Remez, Tal and Nguyen, Tu Anh and Gat, Itai and CONNEAU, Alexis and Kreuk, Felix and Copet, Jade and Defossez, Alexandre and Synnaeve, Gabriel and Dupoux, Emmanuel and Schwartz, Roy and Adi, Yossi},
 booktitle = {Advances in Neural Information Processing Systems},
 pages = {63483--63501},
 title = {{Textually Pretrained Speech Language Models}},
 volume = {36},
 year = {2023}
}

@article{lakhotia2021generative,
  title={{On generative spoken language modeling from raw audio}},
  author={Lakhotia, Kushal and Kharitonov, Eugene and Hsu, Wei-Ning and Adi, Yossi and Polyak, Adam and Bolte, Benjamin and Nguyen, Tu-Anh and Copet, Jade and Baevski, Alexei and Mohamed, Abdelrahman and others},
  journal={Transactions of the Association for Computational Linguistics},
  volume={9},
  pages={1336--1354},
  year={2021}
}

@inproceedings{wang2023minisuperb,
  title={{MiniSuPEBR: Lightweight benchmark for self-supervised speech models}},
  author={Wang, Yu-Hsiang and Chen, Huang-Yu and Chang, Kai-Wei and Hsu, Winston and Lee, Hung-yi},
  booktitle={2023 IEEE Automatic Speech Recognition and Understanding Workshop (ASRU)},
  pages={1--8},
  year={2023}
}

@inproceedings{shi23g_interspeech,
  title     = {{ML-SUPERB: Multilingual Speech Universal PERformance Benchmark}},
  author    = {Jiatong Shi and Dan Berrebbi and William Chen and En-Pei Hu and Wei-Ping Huang and Ho-Lam Chung and Xuankai Chang and Shang-Wen Li and Abdelrahman Mohamed and Hung-yi Lee and Shinji Watanabe},
  year      = {2023},
  booktitle = {Interspeech},
  pages     = {884--888}
}

@inproceedings{shi2024multiresolution,
    title={{Multi-resolution HuBERT: Multi-resolution Speech Self-Supervised Learning with Masked Unit Prediction}},
    author={Jiatong Shi and Hirofumi Inaguma and Xutai Ma and Ilia Kulikov and Anna Sun},
    booktitle={The Twelfth International Conference on Learning Representations},
    year={2024}
}

@inproceedings{shi2023findings,
  title={{Findings of the 2023 ML-SUPERB challenge: Pre-training and evaluation over more languages and beyond}},
  author={Shi, Jiatong and Chen, William and Berrebbi, Dan and Wang, Hsiu-Hsuan and Huang, Wei-Ping and Hu, En-Pei and Chuang, Ho-Lam and Chang, Xuankai and Tang, Yuxun and Li, Shang-Wen and others},
  booktitle={2023 IEEE Automatic Speech Recognition and Understanding Workshop (ASRU)},
  pages={1--8},
  year={2023},
  organization={IEEE}
}

@inproceedings{wang2023_interspeech,
  title     = {{Task- Agnostic Structured Pruning of Speech Representation Models}},
  author    = {Wang, Haoyu and Wang, Siyuan and Zhang, Wei-Qiang and Suo, Hongbin and Wan, Yulong},
  year      = {2023},
  booktitle = {Interspeech},
  pages     = {231–235}
}

@inproceedings{koluguri2022titanet,
  title={{Titanet: Neural model for speaker representation with 1d depth-wise separable convolutions and global context}},
  author={Koluguri, Nithin Rao and Park, Taejin and Ginsburg, Boris},
  booktitle={Proceedings of the IEEE International Conference on Acoustics, Speech, and Signal Processing (ICASSP)},
  pages={8102--8106},
  year={2022},
  organization={IEEE}
}

@article{desplanques2020ecapa,
  title={{ECAPA-TDNN: Emphasized Channel Attention, Propagation and Aggregation in TDNN Based Speaker Verification}},
  author={Desplanques, Brecht and Thienpondt, Jenthe and Demuynck, Kris},
  journal={arXiv preprint arXiv:2005.07143},
  year={2020}
}

@inproceedings{DBLP:conf/icml/ZhuL0W0W24,
  author={Lianghui Zhu and Bencheng Liao and Qian Zhang and Xinlong Wang and Wenyu Liu and Xinggang Wang},
  title={{Vision Mamba: Efficient Visual Representation Learning with Bidirectional State Space Model}},
  year={2024},
  booktitle={International Conference on Machine Learning (ICML)}
}

@INPROCEEDINGS{10890199,
  author={Huang, Hsiang-Wei and Yang, Cheng-Yen and Chai, Wenhao and Jiang, Zhongyu and Hwang, Jeng-Neng},
  booktitle={Proceedings of the IEEE International Conference on Acoustics, Speech, and Signal Processing (ICASSP)}, 
  title={{MambaMOT: State-Space Model as Motion Predictor for Multi-Object Tracking}}, 
  year={2025},
  pages={1-5}
}

@inproceedings{lin2024property,
  title     = {{Property Neurons in Self-Supervised Speech Transformers}},
  author    = {Tzu-Quan Lin and Guan-Ting Lin and Hung-yi Lee and Hao Tang},
  booktitle = {Proceedings of the 2024 IEEE Spoken Language Technology Workshop (SLT)},
  pages     = {401--408},
  year      = {2024},
  publisher = {IEEE}
}

@inproceedings{lenz2025jamba,
  title={{Jamba: Hybrid Transformer-Mamba Language Models}},
  author={Lenz, Barak and others},
  booktitle={The Thirteenth International Conference on Learning Representations},
  year={2025}
}

@article{gu2024mambalineartimesequencemodeling,
  title = {{Mamba: Linear-Time Sequence Modeling with Selective State Spaces}},
  author = {Gu, Albert and Dao, Tri},
  journal = {arXiv preprint arXiv:2312.00752},
  year = {2023}
}

@article{lin2025speechft,
  title   = {{Speech-FT: A Fine-tuning Strategy for Enhancing Speech Representation Models Without Compromising Generalization Ability}},
  author  = {Lin, Tzu-Quan and Huang, Wei-Ping and Tang, Hao and Lee, Hung-yi},
  journal = {arXiv preprint arXiv:2502.12672},
  year    = {2025}
}

@inproceedings{jiang2024dualpathmambashortlongterm,
  title        = {{Dual-path Mamba: Short and Long-term Bidirectional Selective Structured State Space Models for Speech Separation}},
  author       = {Jiang, Xilin and Han, Cong and Mesgarani, Nima},
  booktitle    = {Proceedings of the IEEE International Conference on Acoustics, Speech and Signal Processing (ICASSP)},
  pages={1-5},
  year         = {2025}
}

@article{li2024spmambastatespacemodelneed,
  title        = {{SPMamba: State-space model is all you need in speech separation}},
  author       = {Li, Kai and Chen, Guo and Yang, Runxuan and Hu, Xiaolin},
  journal      = {arXiv preprint arXiv:2404.02063},
  year         = {2024}

}

@inproceedings{lin2024daisy,
  title     = {{DAISY: Data Adaptive Self-Supervised Early Exit for Speech Representation Models}},
  author    = {Lin, Tzu-Quan and Lee, Hung-yi and Tang, Hao},
  booktitle = {Interspeech 2024},
  pages     = {4513--4517},
  year      = {2024}
}

@inproceedings{baevski2020wav2vec20frameworkselfsupervised,
  title        = {{wav2vec 2.0: A Framework for Self-Supervised Learning of Speech Representations}},
  author       = {Baevski, Alexei and Zhou, Yuhao and Mohamed, Abdelrahman and Auli, Michael},
  booktitle    = {Advances in Neural Information Processing Systems},
  volume       = {33},
  pages        = {12449--12460},
  year         = {2020}
}

@INPROCEEDINGS{10832332,
  author={Chao, Rong and Cheng, Wen-Huang and Quatra, Moreno La and Siniscalchi, Sabato Marco and Yang, Chao-Han Huck and Fu, Szu-Wei and Tsao, Yu},
  booktitle={2024 IEEE Spoken Language Technology Workshop (SLT)}, 
  title={{An Investigation of Incorporating Mamba For Speech Enhancement}}, 
  year={2024},
  pages={302-308}
}

@INPROCEEDINGS{10889111,
  author={Zhang, Xiangyu and Ma, Jianbo and Shahin, Mostafa and Ahmed, Beena and Epps, Julien},
  booktitle={Proceedings of the IEEE International Conference on Acoustics, Speech and Signal Processing (ICASSP)}, 
  title={{Rethinking Mamba in Speech Processing by Self-Supervised Models}}, 
  year={2025},
  pages={1-5}
}

@article{zhang2025mambaspeechalternativeselfattention,
  title        = {{Mamba in Speech: Towards an Alternative to Self-Attention}},
  author       = {Zhang, Xiangyu and Zhang, Qiquan and Liu, Hexin and Xiao, Tianyi and Qian, Xinyuan and Ahmed, Beena and Ambikairajah, Eliathamby and Li, Haizhou and Epps, Julien},
  journal      = {arXiv preprint arXiv:2405.12609},
  year         = {2024}
}

@INPROCEEDINGS{gao2024speech,
  author={Gao, Xiaoxue and Chen, Nancy F.},
  booktitle={2024 IEEE Spoken Language Technology Workshop (SLT)}, 
  title={{Speech-Mamba: Long-Context Speech Recognition with Selective State Spaces Models}}, 
  year={2024},
  pages={1-8}
}

@article{pasad2024self,
  title={What do self-supervised speech models know about words?},
  author={Pasad, Ankita and Chien, Chung-Ming and Settle, Shane and Livescu, Karen},
  journal={Transactions of the Association for Computational Linguistics},
  volume={12},
  pages={372--391},
  year={2024}
}

@article{paszke2019pytorch,
  title={{PyTorch: An Imperative Style, High-Performance Deep Learning Library}},
  author={Paszke, Adam and Gross, Sam and Massa, Francisco and Lerer, Adam and Bradbury, James and Chanan, Gregory and Killeen, Trevor and Lin, Zeming and Gimelshein, Natalia and Antiga, Luca and others},
  journal={Advances in Neural Information Processing Systems},
  volume={32},
  year={2019}
}

@inproceedings{NIPS2017_3f5ee243,
 author = {Vaswani, Ashish and Shazeer, Noam and Parmar, Niki and Uszkoreit, Jakob and Jones, Llion and Gomez, Aidan N and Kaiser, \L ukasz and Polosukhin, Illia},
 booktitle = {Advances in Neural Information Processing Systems},
 editor = {I. Guyon and U. Von Luxburg and S. Bengio and H. Wallach and R. Fergus and S. Vishwanathan and R. Garnett},
 pages = {},
 publisher = {Curran Associates, Inc.},
 title = {Attention is All you Need},
 volume = {30},
 year = {2017}
}

@inproceedings{devlin-etal-2019-bert,
    title = "{BERT}: Pre-training of Deep Bidirectional Transformers for Language Understanding",
    author = "Devlin, Jacob  and
      Chang, Ming-Wei  and
      Lee, Kenton  and
      Toutanova, Kristina",
    editor = "Burstein, Jill  and
      Doran, Christy  and
      Solorio, Thamar",
    booktitle = "Proceedings of the 2019 Conference of the North {A}merican Chapter of the Association for Computational Linguistics: Human Language Technologies, Volume 1 (Long and Short Papers)",
    month = jun,
    year = "2019",
    address = "Minneapolis, Minnesota",
    publisher = "Association for Computational Linguistics",
    pages = "4171--4186"
}

@inproceedings{dosovitskiy2021an,
title={An Image is Worth 16x16 Words: Transformers for Image Recognition at Scale},
author={Alexey Dosovitskiy and Lucas Beyer and Alexander Kolesnikov and Dirk Weissenborn and Xiaohua Zhai and Thomas Unterthiner and Mostafa Dehghani and Matthias Minderer and Georg Heigold and Sylvain Gelly and Jakob Uszkoreit and Neil Houlsby},
booktitle={International Conference on Learning Representations},
year={2021}
}

@misc{radford2022robustspeechrecognitionlargescale,
      title={Robust Speech Recognition via Large-Scale Weak Supervision}, 
      author={Alec Radford and Jong Wook Kim and Tao Xu and Greg Brockman and Christine McLeavey and Ilya Sutskever},
      year={2022},
      eprint={2212.04356},
      archivePrefix={arXiv},
      primaryClass={eess.AS}
}

@misc{openai2024gpt4technicalreport,
      title={GPT-4 Technical Report}, 
      author={OpenAI and Josh Achiam and Steven Adler and Sandhini Agarwal and Lama Ahmad and Ilge Akkaya and Florencia Leoni Aleman and Diogo Almeida and Janko Altenschmidt and Sam Altman and Shyamal Anadkat and Red Avila and Igor Babuschkin and Suchir Balaji and Valerie Balcom and Paul Baltescu and Haiming Bao and Mohammad Bavarian and Jeff Belgum and Irwan Bello and Jake Berdine and Gabriel Bernadett-Shapiro and Christopher Berner and Lenny Bogdonoff and Oleg Boiko and Madelaine Boyd and Anna-Luisa Brakman and Greg Brockman and Tim Brooks and Miles Brundage and Kevin Button and Trevor Cai and Rosie Campbell and Andrew Cann and Brittany Carey and Chelsea Carlson and Rory Carmichael and Brooke Chan and Che Chang and Fotis Chantzis and Derek Chen and Sully Chen and Ruby Chen and Jason Chen and Mark Chen and Ben Chess and Chester Cho and Casey Chu and Hyung Won Chung and Dave Cummings and Jeremiah Currier and Yunxing Dai and Cory Decareaux and Thomas Degry and Noah Deutsch and Damien Deville and Arka Dhar and David Dohan and Steve Dowling and Sheila Dunning and Adrien Ecoffet and Atty Eleti and Tyna Eloundou and David Farhi and Liam Fedus and Niko Felix and Simón Posada Fishman and Juston Forte and Isabella Fulford and Leo Gao and Elie Georges and Christian Gibson and Vik Goel and Tarun Gogineni and Gabriel Goh and Rapha Gontijo-Lopes and Jonathan Gordon and Morgan Grafstein and Scott Gray and Ryan Greene and Joshua Gross and Shixiang Shane Gu and Yufei Guo and Chris Hallacy and Jesse Han and Jeff Harris and Yuchen He and Mike Heaton and Johannes Heidecke and Chris Hesse and Alan Hickey and Wade Hickey and Peter Hoeschele and Brandon Houghton and Kenny Hsu and Shengli Hu and Xin Hu and Joost Huizinga and Shantanu Jain and Shawn Jain and Joanne Jang and Angela Jiang and Roger Jiang and Haozhun Jin and Denny Jin and Shino Jomoto and Billie Jonn and Heewoo Jun and Tomer Kaftan and Łukasz Kaiser and Ali Kamali and Ingmar Kanitscheider and Nitish Shirish Keskar and Tabarak Khan and Logan Kilpatrick and Jong Wook Kim and Christina Kim and Yongjik Kim and Jan Hendrik Kirchner and Jamie Kiros and Matt Knight and Daniel Kokotajlo and Łukasz Kondraciuk and Andrew Kondrich and Aris Konstantinidis and Kyle Kosic and Gretchen Krueger and Vishal Kuo and Michael Lampe and Ikai Lan and Teddy Lee and Jan Leike and Jade Leung and Daniel Levy and Chak Ming Li and Rachel Lim and Molly Lin and Stephanie Lin and Mateusz Litwin and Theresa Lopez and Ryan Lowe and Patricia Lue and Anna Makanju and Kim Malfacini and Sam Manning and Todor Markov and Yaniv Markovski and Bianca Martin and Katie Mayer and Andrew Mayne and Bob McGrew and Scott Mayer McKinney and Christine McLeavey and Paul McMillan and Jake McNeil and David Medina and Aalok Mehta and Jacob Menick and Luke Metz and Andrey Mishchenko and Pamela Mishkin and Vinnie Monaco and Evan Morikawa and Daniel Mossing and Tong Mu and Mira Murati and Oleg Murk and David Mély and Ashvin Nair and Reiichiro Nakano and Rajeev Nayak and Arvind Neelakantan and Richard Ngo and Hyeonwoo Noh and Long Ouyang and Cullen O'Keefe and Jakub Pachocki and Alex Paino and Joe Palermo and Ashley Pantuliano and Giambattista Parascandolo and Joel Parish and Emy Parparita and Alex Passos and Mikhail Pavlov and Andrew Peng and Adam Perelman and Filipe de Avila Belbute Peres and Michael Petrov and Henrique Ponde de Oliveira Pinto and Michael and Pokorny and Michelle Pokrass and Vitchyr H. Pong and Tolly Powell and Alethea Power and Boris Power and Elizabeth Proehl and Raul Puri and Alec Radford and Jack Rae and Aditya Ramesh and Cameron Raymond and Francis Real and Kendra Rimbach and Carl Ross and Bob Rotsted and Henri Roussez and Nick Ryder and Mario Saltarelli and Ted Sanders and Shibani Santurkar and Girish Sastry and Heather Schmidt and David Schnurr and John Schulman and Daniel Selsam and Kyla Sheppard and Toki Sherbakov and Jessica Shieh and Sarah Shoker and Pranav Shyam and Szymon Sidor and Eric Sigler and Maddie Simens and Jordan Sitkin and Katarina Slama and Ian Sohl and Benjamin Sokolowsky and Yang Song and Natalie Staudacher and Felipe Petroski Such and Natalie Summers and Ilya Sutskever and Jie Tang and Nikolas Tezak and Madeleine B. Thompson and Phil Tillet and Amin Tootoonchian and Elizabeth Tseng and Preston Tuggle and Nick Turley and Jerry Tworek and Juan Felipe Cerón Uribe and Andrea Vallone and Arun Vijayvergiya and Chelsea Voss and Carroll Wainwright and Justin Jay Wang and Alvin Wang and Ben Wang and Jonathan Ward and Jason Wei and CJ Weinmann and Akila Welihinda and Peter Welinder and Jiayi Weng and Lilian Weng and Matt Wiethoff and Dave Willner and Clemens Winter and Samuel Wolrich and Hannah Wong and Lauren Workman and Sherwin Wu and Jeff Wu and Michael Wu and Kai Xiao and Tao Xu and Sarah Yoo and Kevin Yu and Qiming Yuan and Wojciech Zaremba and Rowan Zellers and Chong Zhang and Marvin Zhang and Shengjia Zhao and Tianhao Zheng and Juntang Zhuang and William Zhuk and Barret Zoph},
      year={2024},
      eprint={2303.08774},
      archivePrefix={arXiv},
      primaryClass={cs.CL}
}

@inproceedings{lin2025identifying,
  title        = {{Identifying Speaker Information in Feed-Forward Layers of Self-Supervised Speech Transformers}},
  author       = {Lin, Tzu-Quan and Cheng, Hsi-Chun and Lee, Hung-yi and Tang, Hao},
  booktitle      = {APSIPA ASC 2025},
  year         = {2025}
}

@article{yang2024large,
  title        = {A Large-Scale Evaluation of Speech Foundation Models},
  author       = {Yang, Shu-wen and Chang, Heng-Jui and Huang, Zili and Liu, Andy T. and Lai, Cheng-I and Wu, Haibin and Shi, Jiatong and Chang, Xuankai and Tsai, Hsiang-Sheng and Huang, Wen-Chin and Feng, Tzu-hsun and Chi, Po-Han and Lin, Yist Y. and Chuang, Yung-Sung and Huang, Tzu-Hsien and Tseng, Wei-Cheng and Lakhotia, Kushal and Li, Shang-Wen and Mohamed, Abdelrahman and Watanabe, Shinji and Lee, Hung-yi},
  journal      = {IEEE/ACM Transactions on Audio, Speech, and Language Processing},
  volume       = {32},
  pages        = {2884--2899},
  year         = {2024},
}

@article{suleman2024stablemamba,
  title={StableMamba: Distillation-free Scaling of Large SSMs for Images and Videos},
  author={Suleman, Hamid and Wasim, Syed Talal and Naseer, Muzammal and Gall, Juergen},
  journal={arXiv preprint arXiv:2409.11867},
  year={2024}
}

@inproceedings{shaker2025groupmamba,
  title={GroupMamba: Efficient Group-Based Visual State Space Model},
  author={Shaker, Abdelrahman and Wasim, Syed Talal and Khan, Salman and Khan, Fahad Shahbaz and Anwer, Rao Muhammad},
  booktitle={Proceedings of the IEEE/CVF Conference on Computer Vision and Pattern Recognition (CVPR)},
  year={2025}
}

@article{patro2024simba,
  title={SiMBA: Simplified Mamba-based Architecture for Vision and Multivariate Time Series},
  author={Patro, Badri N and Agneeswaran, Vijay S},
  journal={arXiv preprint arXiv:2403.15360},
  year={2024}
}
\appendix

\section{Architectural Differences between ExtBiMamba and InnBiMamba}
\label{sec:bimamba_appendix}

\begin{figure}[ht]
    \centering
    \includegraphics[scale=0.45]{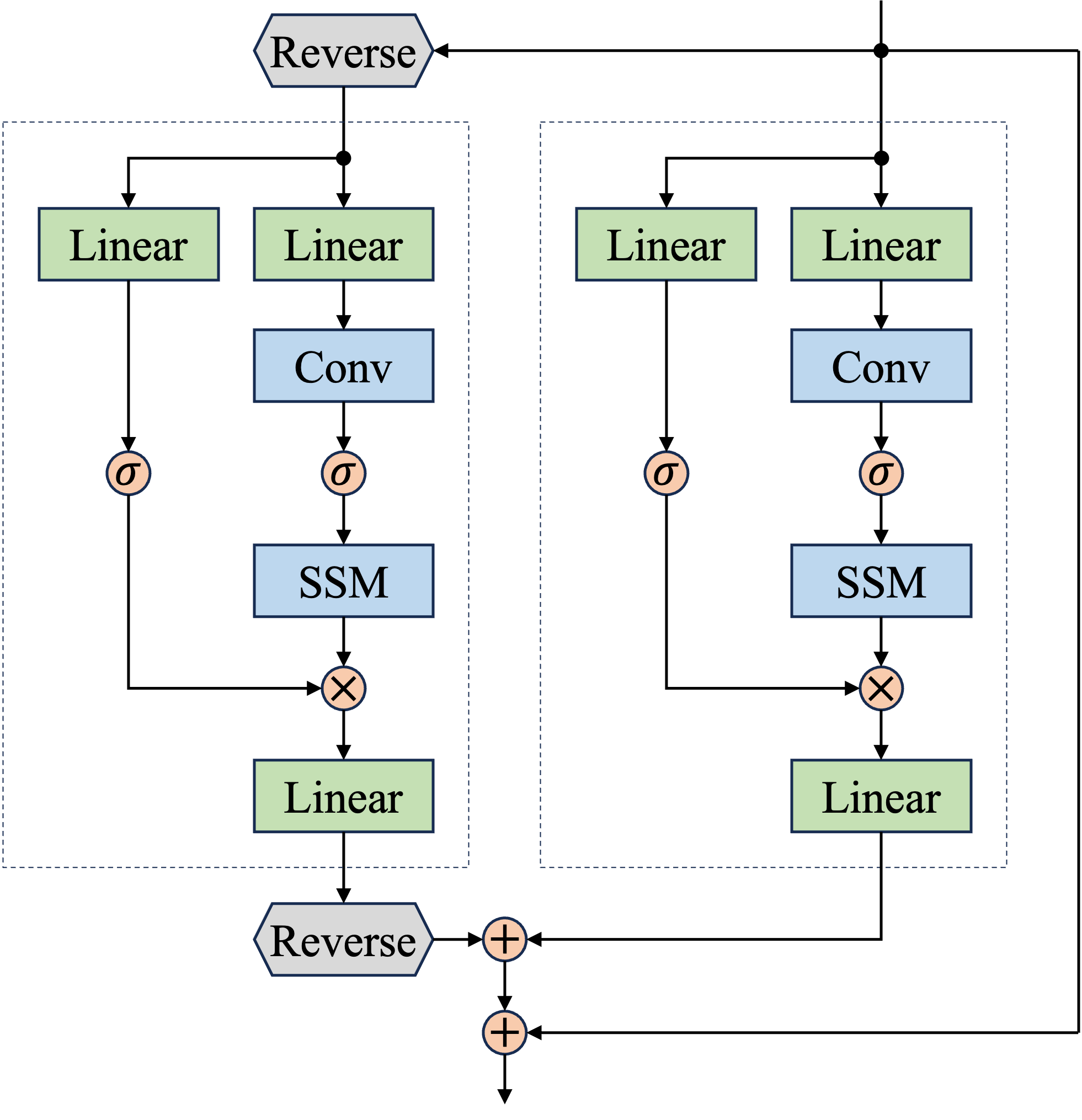}
    \caption{Architecture of ExtBiMamba, illustrating its independent forward and backward branches.}
    \label{fig:extbimamba_appendix}
\end{figure}

\begin{figure}[ht]
    \centering
    \includegraphics[scale=0.45]{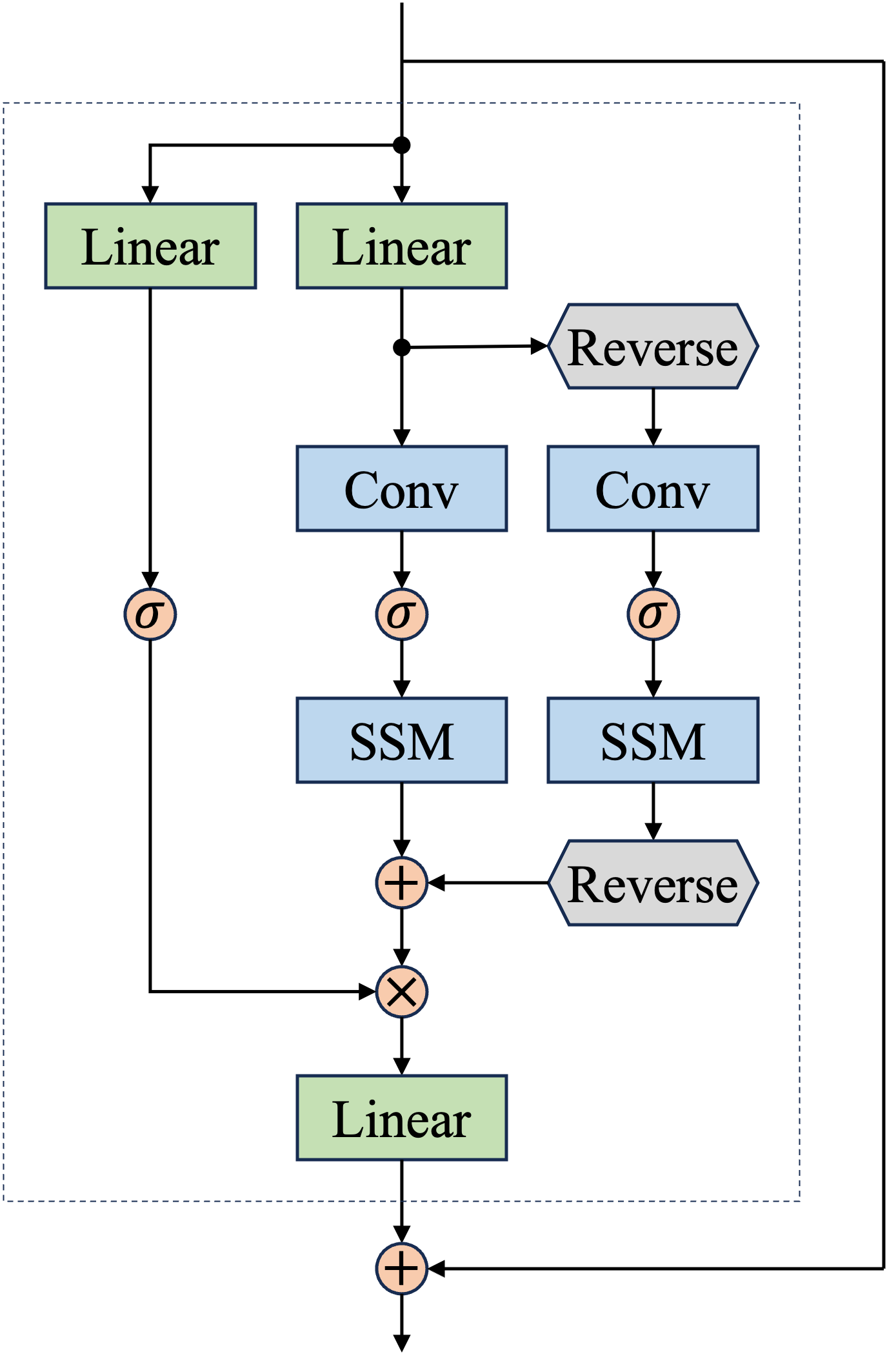}
    \caption{Architecture of InnBiMamba, where the forward and backward branches share projections but duplicate the convolutional layer and SSM module.}
    \label{fig:innbimamba_appendix}
\end{figure}

Figure~\ref{fig:extbimamba_appendix} and ~\ref{fig:innbimamba_appendix} illustrate the architectures of ExtBiMamba and InnBiMamba to clarify their differences.

\section{Additional Results for Phone Purity of Quantized Representations}
\label{sec:phone_purity_additional}

\begin{figure}[ht]
  \centering
  
  \begin{subfigure}{\linewidth}
    \centering
    \begin{tikzpicture}
          \definecolor{MyOrange}{HTML}{FF7F0E}
          \definecolor{MyBlue}{HTML}{1F77B4}
          \definecolor{MyLightOrange}{HTML}{FFBF86}
          \definecolor{MyLightBlue}{HTML}{8FBBD9}

          \draw[thick, MyOrange] (0,0) -- +(0.8,0);
          \node[MyOrange] at (0.4,0) {\scalebox{2}{\pgfuseplotmark{triangle*}}};
          \node[anchor=west] at (1.0,0) {Mamba};

          \draw[thick, MyBlue] (3.5,0) -- +(0.8,0);
          \node[MyBlue] at (3.9,0) {\scalebox{1.5}{\pgfuseplotmark*}};
          \node[anchor=west] at (4.5,0) {Causal Transformer};

          \draw[thick, MyLightOrange] (0,-0.6) -- +(0.8,0);
          \node[MyLightOrange] at (0.4,-0.6) {\scalebox{2}{\pgfuseplotmark{triangle*}}};
          \node[anchor=west] at (1.0,-0.6) {ExtBiMamba};

          \draw[thick, MyLightBlue] (3.5,-0.6) -- +(0.8,0);
          \node[MyLightBlue] at (3.9,-0.6) {\scalebox{1.5}{\pgfuseplotmark*}};
          \node[anchor=west] at (4.5,-0.6) {Transformer};
    \end{tikzpicture}
  \end{subfigure}

  \vspace{0.5em}

  \begin{subfigure}{1.0\linewidth}
    \centering
    \includegraphics[width=\linewidth]{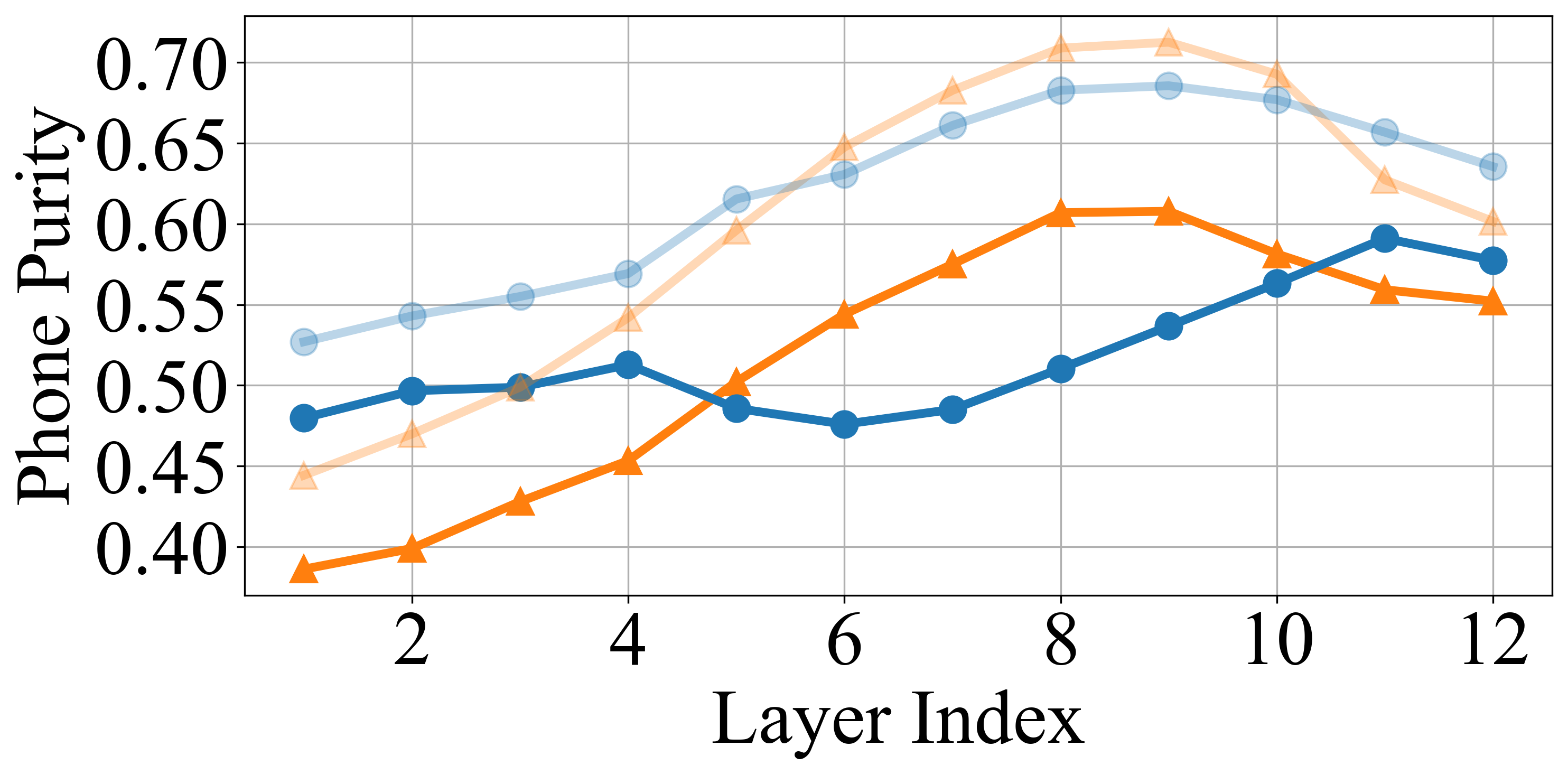}
    \caption*{$k=500$}
  \end{subfigure}
  \begin{subfigure}{1.0\linewidth}
    \centering
    \includegraphics[width=\linewidth]{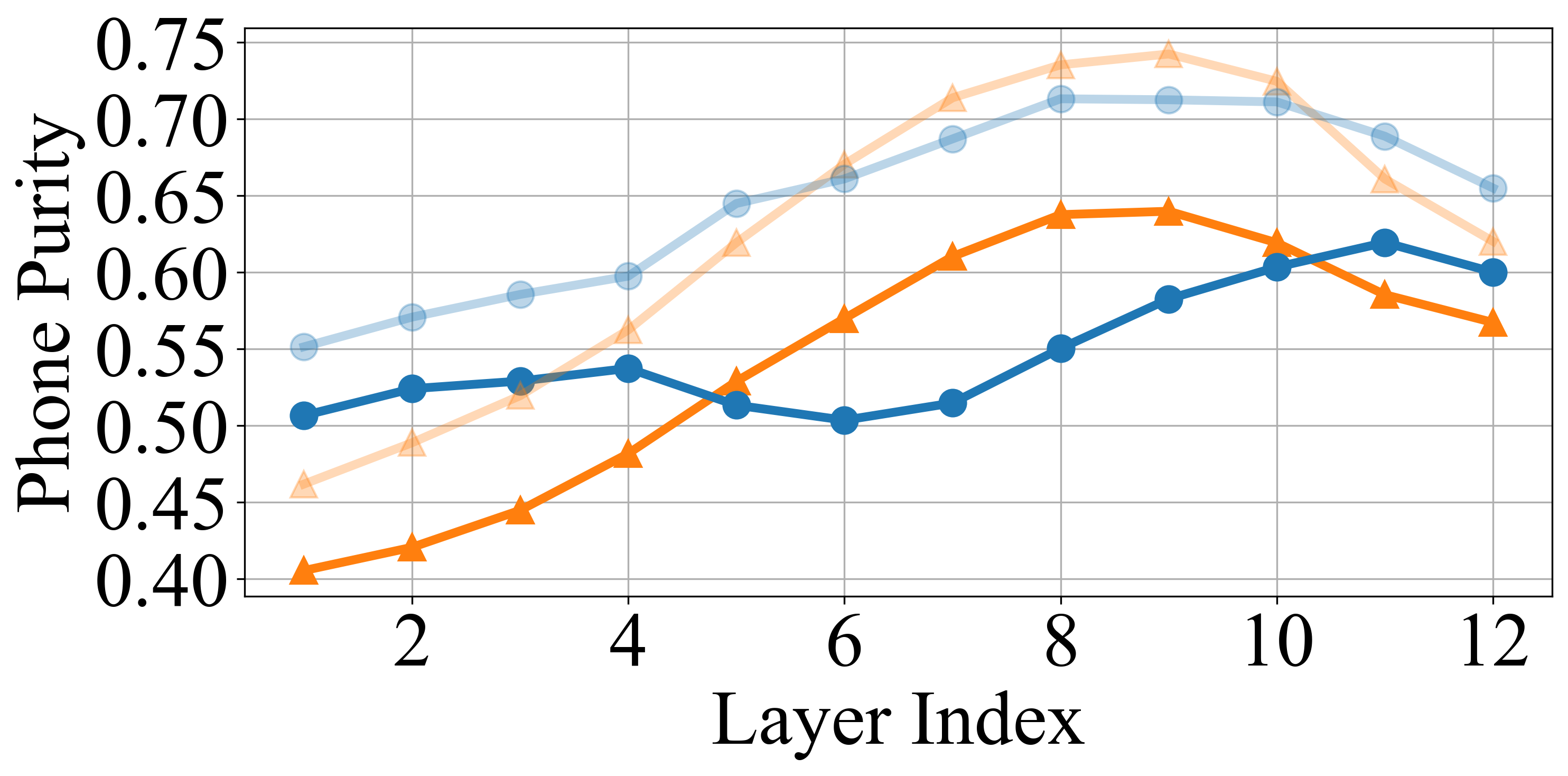}
    \caption*{$k=1000$}
  \end{subfigure}

  \caption{Layer-wise phone purity of HuBERT models with k-means clustering at $k=500$ and $k=1000$, as a supplement to the $k=100$ results in Figure~\ref{fig:phone_purity}.}
  \label{fig:phone_purity_additional}
\end{figure}

Figure~\ref{fig:phone_purity_additional} shows the layer-wise phone purity of HuBERT models with k-means clustering at $k=500$ and $k=1000$, complementing the $k=100$ results in Figure~\ref{fig:phone_purity}.

\section{Computation of overall SUPERB score}
\label{sec:superb-score}
To provide a unified comparison across tasks, we use the overall SUPERB score (SUPERB$_S$)~\citep{feng2023superb}. Each task’s metric is linearly scaled between the performance of baseline features (FBank) and the state-of-the-art (SOTA) results. This formulation implicitly accounts for task difficulty: when FBank already performs close to SOTA, even small improvements are emphasized more; whereas for tasks with a large gap between FBank and SOTA, the same improvement is considered less significant. For tasks with multiple metrics, scores are first averaged within the task, then across all tasks. The final result is multiplied by 1000 for readability. In this work, we follow the default SUPERB evaluation protocol and compute SUPERB$_S$ on four representative tasks: phoneme recognition (PR), speaker identification (SID), emotion recognition (ER), and slot filling (SF). The formulation is:

\begin{equation}
\Phi_{t,j}(f) = 
\frac{\phi_{t,j}(f) - \phi_{t,j}(\text{FBank})}
     {\phi_{t,j}(\text{SOTA}) - \phi_{t,j}(\text{FBank})}
\end{equation}

\begin{equation}
\text{SUPERB}_S(f) = 
\frac{1000}{|T|} \sum_{t \in T} \frac{1}{|M_t|} \sum_{j \in M_t} \Phi_{t,j}(f)
\end{equation}

Here, $\phi_{t,j}(f)$ denotes the score of model $f$ on metric $j$ of task $t$, $T$ is the set of tasks, and $M_t$ is the set of metrics for task $t$.
The results of FBank and SOTA are taken from the SUPERB journal extension~\citep{yang2024large}.


\section{Training Stability Analysis}

\label{sec:training-stability}

\begin{figure*}[ht]
    \centering
    \includegraphics[scale=0.5]{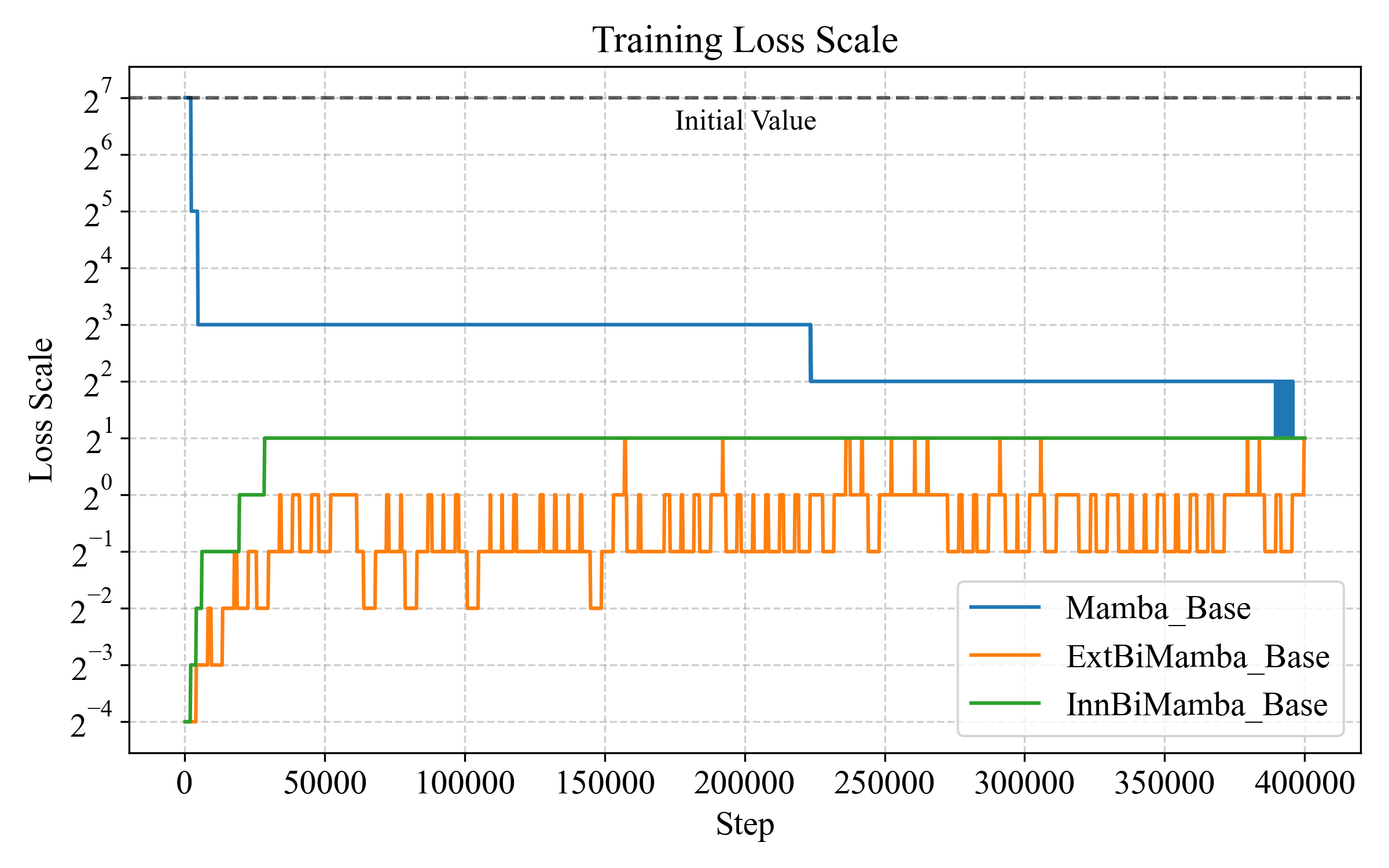}
    \caption{Training dynamics of the loss scale for different Mamba variants.}
    \label{fig:training_stability}
\end{figure*}

We visualize the variation of loss scale at each training step for different Mamba variants in Figure~\ref{fig:training_stability}. These variants—Mamba, InnBiMamba, and ExtBiMamba—represent a progression of increasing architectural complexity. In our training pipeline, the loss scale is halved whenever a gradient overflow occurs and doubled when no overflow is detected for several consecutive steps.

As shown in the figure, the vanilla Mamba model rarely overflows and stabilizes immediately. InnBiMamba overflows frequently during early training but gradually converges to stable loss scales. In contrast, ExtBiMamba, the most complex architecture among the three, exhibits persistent oscillations throughout pre-training, indicating continual instability.

This instability aligns with observations in prior work on Mamba-based visual backbones~\citep{suleman2024stablemamba, shaker2025groupmamba, patro2024simba}, where scaling up often leads to unstable training compared to Transformers. To ensure stability in larger-scale regimes with increased batch sizes or diverse corpora, future designs may require auxiliary mechanisms such as hybrid blocks that incorporate attention mechanisms or distillation strategies.

\begin{table*}[ht]
    \centering
    \caption{Full SUPERB evaluation results for Causal Transformer Base, Mamba+MLP Base, Transformer Base, and ExtBiMamba Base. $\text{SUPERB}_{\text{FULL}}$ is computed across all 10 tasks. Note that $\text{SUPERB}_{\text{FULL}}$ is not directly comparable to $\text{SUPERB}_S$ reported elsewhere in the paper.}
    \footnotesize
    \setlength{\tabcolsep}{2.5pt}
    \scalebox{1.0}{
    \renewcommand{\arraystretch}{1.1}
    \begin{tabular}{l|ccccccccccc|c}
    \hline
     \multirow{2}{*}{\textbf{Model}} & \textbf{PR} & \textbf{ASR} & \textbf{KS} & \textbf{QbE} & \textbf{SID} & \textbf{ASV} & \textbf{SD} & \textbf{ER} & \textbf{IC} & \multicolumn{2}{c|}{\textbf{SF}} & \multirow{2}{*}{$\text{SUPERB}_{\text{FULL}}$}\\ 
    & \scriptsize{PER\%$\downarrow$} & \scriptsize{WER\%$\downarrow$} & \scriptsize{ACC\%$\uparrow$} & \scriptsize{MTWV$\uparrow$} & \scriptsize{ACC\%$\uparrow$} & \scriptsize{EER\%$\downarrow$} & \scriptsize{DER\%$\downarrow$} & \scriptsize{ACC\%$\uparrow$} & \scriptsize{ACC\%$\uparrow$} & \scriptsize{F1$\uparrow$} & \scriptsize{CER\%$\downarrow$} & \\
    \hline
    \hline
    Causal Trans. Base & 13.87 & 13.55 & 95.13 & 0.0356 & 60.04 & \textbf{7.66} & 8.23 & \textbf{63.33} & \textbf{94.23} & \textbf{81.91} & \textbf{35.55} & 608.1 \\ 
    Mamba+MLP Base & \textbf{11.72} & \textbf{12.25} & \textbf{95.91} & \textbf{0.0590} & \textbf{73.48} & 7.77 & \textbf{7.36} & 61.72 & 89.62 & 80.99 & 37.12 & \textbf{651.6} \\
    \hline
    Trans. Base & \textbf{7.49} & \textbf{9.16} & 95.01 & \textbf{0.0792} & \textbf{75.77} & \textbf{5.91} & 6.86 & \textbf{62.36} & \textbf{97.44} & \textbf{87.18} & \textbf{26.38} & \textbf{772.4} \\
    ExtBiMamba Base & 10.65 & 11.41 & \textbf{96.36} & 0.0586 & 68.31 & 7.07 & \textbf{6.76} & 61.24 & 91.70 & 83.14 & 33.89 & 683.5 \\ 
    \hline
    \end{tabular}
    }
    \label{tab:full-superb-appendix}
\end{table*}

\section{Full SUPERB Evaluation}
We provide the full SUPERB benchmark evaluation results for representative models to verify whether the trends observed in Section~\ref{sec:comparative_performance}, based on the four-task subset ($SUPERB_{S}$), hold across the complete set of tasks. The detailed scores across all 10 tasks are summarized in Table~\ref{tab:full-superb-appendix}. $\text{SUPERB}_{\text{FULL}}$ denotes an overall SUPERB score computed across the 10 tasks. Note that $\text{SUPERB}_{\text{FULL}}$ is not directly comparable to $\text{SUPERB}_S$ reported elsewhere in the paper.

\subsection{Causal Setting}
The full evaluation reinforces our observation that Mamba-based architectures are well-suited for causal speech self-supervised modeling:
\begin{itemize}
    \item Performance Advantage: Mamba+MLP Base achieves a significantly higher $\text{SUPERB}_{\text{FULL}}$ score of 651.6 compared to 608.1 for the Causal Transformer Base, demonstrating superior overall representation quality.
    \item Task-Specific Strengths: Mamba demonstrates clear superiority in Content-related tasks (PR, ASR, KS, QbE) and Speaker-related tasks (SID, SD).
    \item Limitations: Mamba exhibits a performance gap in Paralinguistic (ER) and Semantic tasks (IC, SF) compared to the Causal Transformer.
\end{itemize}

\subsection{Bidirectional Setting}
In the bidirectional setting, the results confirm that the Transformer remains the dominant architecture at the Base scale:
\begin{itemize}
    \item Transformer Superiority: The standard Transformer Base outperforms ExtBiMamba Base across nearly all task categories, achieving a $\text{SUPERB}_{\text{FULL}}$ score of 772.4 versus ExtBiMamba’s 683.59.
    \item Scalability Challenges: These results support our finding that while Mamba is a powerful alternative in causal settings, its scalability in bidirectional settings remains a challenge.
\end{itemize}

\end{document}